%% file: main.tex
\documentclass[10pt,twocolumn,letterpaper]{article}

\usepackage[pagenumbers]{wacv} 

\usepackage{graphicx}
\usepackage{amsmath}
\usepackage{amssymb}
\usepackage{booktabs}
\usepackage[font={footnotesize}]{caption}
\usepackage[numbers,sort&compress]{natbib}

\usepackage[pagebackref,breaklinks,colorlinks]{hyperref}

\usepackage[capitalize]{cleveref}
\crefname{section}{Sec.}{Secs.}
\Crefname{section}{Section}{Sections}
\Crefname{table}{Table}{Tables}
\crefname{table}{Tab.}{Tabs.}

\def\wacvPaperID{909} 
\def\confName{WACV}
\def\confYear{2025}

\input{utils/head}

\newcommand{\FGSM}{{\texttt{FGSM}}}
\newcommand{\CW}{{\texttt{CW}}}
\newcommand{\PGD}{{\texttt{PGD}}}
\newcommand{\AutoAttack}{{\texttt{AutoAttack}}}
\newcommand{\AutoAttackShort}{{\texttt{AA}}}
\newcommand{\AutoPGD}{{\texttt{AutoPGD}}}
\newcommand{\SquareAttack}{{\texttt{SquareAttack}}}
\newcommand{\SquareAttackShort}{{\texttt{Square}}}
\newcommand{\ZOSignSGD}{{\texttt{ZO-signSGD}}}
\newcommand{\NES}{{\texttt{NES}}} 
 
\newcommand{\CIFAR}{{\texttt{CIFAR-10}}}
\newcommand{\CIFARM}{{\texttt{CIFAR-100}}}
\newcommand{\TImageNet}{{\texttt{Tiny-ImageNet}}}
\newcommand{\KS}{{\texttt{KS}}}
\newcommand{\AF}{{\texttt{AF}}}
\newcommand{\WS}{{\texttt{WS}}}
\newcommand{\AT}{{\texttt{AT}}}

\newcommand{\PEN}{{\text{PEN}}}

\newcommand{\MPN}{{\text{MPN}}}
\newcommand{\Conv}{{\texttt{ConvNet-4}}}
\newcommand{\MLP}{{\texttt{MLP}}}

\newcommand{\VM}{{\text{VM}}}
\newcommand{\RED}{{\text{RED}}}

\usepackage{diagbox}

\definecolor{bluegray}{rgb}{0.4, 0.6, 0.8}
\definecolor{ceruleanblue}{rgb}{0.16, 0.32, 0.75}
\hypersetup{
 colorlinks=true,
 citecolor=ceruleanblue,
 linkcolor=ceruleanblue,
 urlcolor=black}

\begin{document}

\title{Can Adversarial Examples Be Parsed to Reveal Victim Model Information?}


\author{Yuguang Yao$^{1*}$, Jiancheng Liu$^{1*}$, Yifan Gong$^{2*}$, Xiaoming Liu$^1$, Yanzhi Wang$^2$, Xue Lin$^2$, Sijia Liu$^{1,3}$\\
$^1$Michigan State University, ~~$^2$Northeastern University, ~~$^3$MIT-IBM Watson AI Lab\\
}

\maketitle
\def\thefootnote{*}\footnotetext{Equal contributions.}

\begin{abstract}
\input{secs/abstract_SLiu}

\end{abstract}

\input{secs/intro_SLiu}
\input{secs/related_works.tex}
\input{secs/prob_statement_SLiu}
\input{secs/methods_SLiu}

\input{secs/experiments_SLiu}
\input{secs/conclusion.tex}
\input{secs/acknowledgement}

{\small
\bibliographystyle{unsrtnat}
\bibliography{refs_bibs/ref_adv}
}

\input{secs/appendix}

\end{document}

%% file: utils/head.tex
\usepackage[utf8]{inputenc} 
\usepackage[T1]{fontenc}    
\usepackage{url}            
\usepackage{booktabs}       
\usepackage{amsfonts}       
\usepackage{nicefrac}       
\usepackage{microtype}      

 \usepackage{graphicx}
\usepackage{pifont}
\usepackage{adjustbox}
\usepackage{lipsum}
\usepackage{color}
\usepackage{wrapfig}
\usepackage{booktabs}
\usepackage[textsize=scriptsize]{todonotes}
\usepackage{multirow,mathtools} 
\usepackage{threeparttable}

\definecolor{lightblue}{rgb}{0.68, 0.85, 0.9}
\definecolor{lightgreen}{rgb}{0.56, 0.93, 0.56}
\definecolor{lightskyblue}{rgb}{0.53, 0.81, 0.98}
\definecolor{non-photoblue}{rgb}{0.64, 0.87, 0.93}
\definecolor{magicmint}{rgb}{0.67, 0.94, 0.82}
\definecolor{mossgreen}{rgb}{0.68, 0.87, 0.68}
\definecolor{salmon}{rgb}{1.0, 0.55, 0.41}
\definecolor{babypink}{rgb}{0.96, 0.76, 0.76}

\DeclareMathOperator*{\minimize}{\text{minimize}}

\DeclareMathAlphabet\mathbfcal{OMS}{cmsy}{b}{n}

\def\btheta{\boldsymbol{\theta}}
\def\bdelta{\boldsymbol{\delta}}

\def\bpsi{\boldsymbol{\psi}}

\usepackage{color, colortbl}
\definecolor{Gray}{gray}{0.93}
\definecolor{Orange}{rgb}{1,0.5,0}
\definecolor{DGray}{gray}{0.83}
\definecolor{LightCyan}{rgb}{0.88,1,1}

\usepackage[most]{tcolorbox}
\newtcolorbox{mybox}[2][]{%
  attach boxed title to top center
               = {yshift=-8pt},
  colback      = Gray,
  colframe     = black,
  fonttitle    = \bfseries,
  colbacktitle = white,
  title        = #2,#1,
  enhanced,
}

\DeclarePairedDelimiterX{\inp}[2]{\langle}{\rangle}{#1, #2}

\makeatletter
\newcommand*{\rom}[1]{\expandafter\@slowromancap\romannumeral #1@}
\makeatother
\newcommand{\mycomment}[1]{}

\usepackage{xcolor,pifont}
\newcommand*\colourcheck[1]{%
  \expandafter\newcommand\csname #1check\endcsname{\textcolor{#1}{\ding{52}}}%
}
\colourcheck{blue}
\colourcheck{green}
\colourcheck{red}
\newcommand*\colourcross[1]{%
  \expandafter\newcommand\csname #1cross\endcsname{\textcolor{#1}{\ding{56}}}%
}
\colourcross{red}

%% file: secs/abstract_SLiu.tex
Numerous adversarial attack methods have been developed to generate imperceptible image perturbations that can cause erroneous predictions of state-of-the-art machine learning (ML) models, in particular, deep neural networks (DNNs). 
Despite extensive research on adversarial attacks (\textit{a.k.a.}, adversarial examples), little effort was made to uncover  `arcana'  carried in these examples.  
In this study, we explore the feasibility of deducing data-agnostic information about the \textit{victim model} ({\VM})--specifically, the characteristics of the ML model or DNN targeted for attack generation--from data-specific adversarial examples. 
We refer to this process as \textit{model parsing of adversarial attacks}.
We approach the model parsing problem as a supervised learning task, aiming to attribute categories of {\VM} characteristics (including architecture type, kernel size, activation function, and weight sparsity) to individual adversarial examples. To this end, we have assembled a model parsing dataset featuring adversarial attacks spanning $7$ distinct types, sourced from $135$ victim models. These models are systematically varied across $5$ architecture types, $3$ kernel size configurations, $3$ activation function categories, and $3$ levels of weight sparsity ratios.
We demonstrate that a simple, supervised model parsing network ({\MPN}) is possible to uncover concealed details of the {\VM} from adversarial examples. 
We also validate the practicality of model parsing from adversarial attacks by examining the effects of diverse training and evaluation factors on parsing performance. This includes investigating the impact of the formats of input attacks used for parsing and assessing generalization capabilities in out-of-distribution scenarios. Furthermore, we show how the proposed {\MPN} can reveal the source {\VM} attributes in transfer attacks, shedding light on a potential connection between model parsing and   attack transferability.
Code is available at \url{https://github.com/OPTML-Group/RED-adv}.
 %
%

%% file: secs/intro_SLiu.tex
\section{Introduction}

\begin{figure}[tb]
\centerline{
\includegraphics[width=0.49\textwidth]{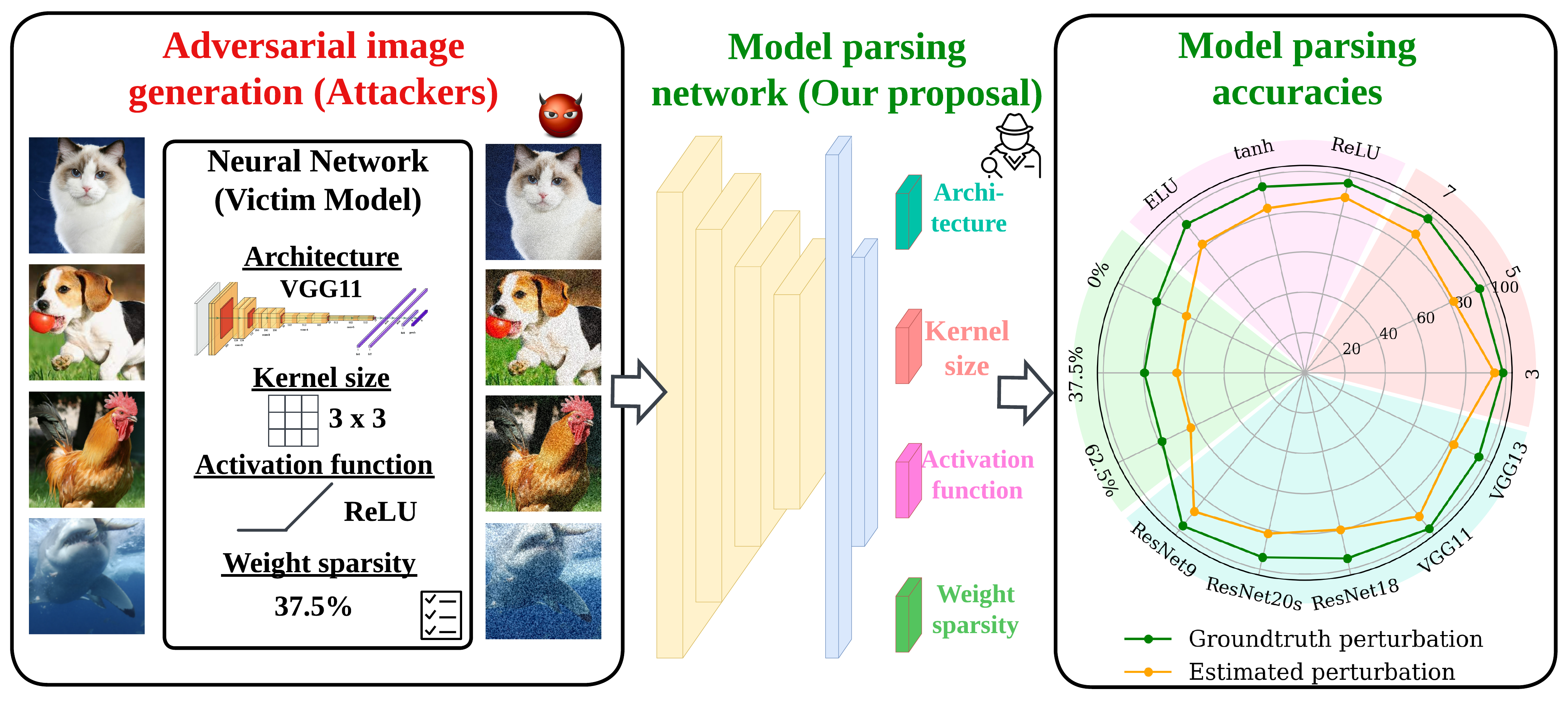}
}
\caption{\footnotesize{
Schematic overview of model parsing from adversarial attacks. (Left) Attack generation leveraging the {\VM} (victim model), with model attributes including architecture type, kernel size, activation function, and weight sparsity. (Middle) Proposed model parsing network ({\MPN}), aiming to classify {\VM} attributes based on adversarial examples.  (Right)  Demonstrating the efficacy of {\MPN} in accurately parsing model attributes from PGD attacks \cite{madry2017towards} against ResNet9 on {\CIFAR}. Performance metrics for {\MPN} are showcased across two distinct types of input:  actual adversarial perturbations and estimated adversarial perturbations, detailed in Sec.\,\ref{sec: Methods}.
}
}
\label{fig: overview}
\end{figure}

Adversarial attacks, in terms of tiny (imperceptible) input perturbations crafted to \textit{fool} the decisions of machine learning (\textbf{ML}) models, have emerged as a primary   security concern of ML in  a wide range of  vision applications\cite{szegedy2013intriguing,goodfellow2014explaining,xie2017adversarial,tu2020physically,antun2020instabilities,raj2020improving,jiang2020robust,fan2021does,bommasani2021opportunities,maus2023adversarial}.
Therefore, a vast amount of prior works have been devoted to answering the questions of    \textit{how to generate} adversarial attacks for adversarial robustness evaluation  {\cite{goodfellow2014explaining,madry2017towards,carlini2017towards,croce2020reliable,chen2017zoo,liu2018signsgd,ilyas2018black,andriushchenko2020square,xie2019improving, xiao2018spatially,moosavi2016deepfool, brendel2017decision}} and  \textit{how to defend} against these attacks for robustness enhancement \cite{madry2017towards,zhang2019theoretically,wong2017provable,salman2020denoised,wong2020fast,carmon2019unlabeled,shafahi2019adversarial,zhou2022trace,grosse2017statistical,yang2020ml,metzen2017detecting,meng2017magnet,wojcik2020adversarial,shi2021online,yoon2021adversarial,srinivasan2021robustifying,zhang2022revisiting,zhang2022robustify}. These two questions are also closely interrelated, with insights from one contributing to the understanding of the other.

In the plane of attack generation, a variety of attack methods have been developed, ranging from gradient-based 
 (white-box, {perfect-knowledge)} attacks  \cite{goodfellow2014explaining,moosavi2016deepfool, madry2017towards,carlini2017towards,xie2019improving,croce2020reliable}  to   query-based (black-box, {restricted-knowledge}) attacks 
\cite{brendel2017decision,chen2017zoo,liu2018signsgd,ilyas2018black,andriushchenko2020square}. Understanding the attack generation process allows us to further understand attacks' characteristics and their specialties. For example, unlike Deepfake images that are created using generative models   \cite{wang2020cnn,asnani2021reverse,dhariwal2021diffusion, yu2019attributing, frank2020leveraging, guarnera2020deepfake, dzanic2020fourier},
 adversarial examples are typically generated through a distinct process involving \textbf{(a)} a simple, deterministic perturbation optimizer (\textit{e.g.}, fast gradient sign method  in \cite{goodfellow2014explaining}), \textbf{(b)} a specific input example (\textit{e.g.}, an image),   and \textbf{(c)} 
a targeted, well-trained victim model (\textbf{\VM}), \textit{i.e.}, an ML model that the adversary aims to compromise. 
In this context, both (a) and (b) interact with and depend on the {\VM} for the generation of attacks. 
The creation of adversarial examples also plays a pivotal role in advancing the development of adversarial defenses, such as
robust training \cite{madry2017towards,zhang2019theoretically,wong2017provable,salman2020denoised,wong2020fast,carmon2019unlabeled,shafahi2019adversarial,zhang2022revisiting}, adversarial detection {\cite{zhou2022trace,grosse2017statistical,yang2020ml,metzen2017detecting,meng2017magnet,wojcik2020adversarial,liao_defense_2018}},  and  adversarial purification  {\cite{srinivasan2021robustifying,shi2021online,yoon2021adversarial,nie2022diffusion}}. 


Beyond traditional attack generation and defensive strategies, recent research \cite{nicholson2023reverse,gong2022reverse,wang2023can,goebel2021attribution,souri2021identification,thaker2022reverse,guo2023scalable, maini2021perturbation,zhou2022trace} has begun to explore and analyze adversarial attacks within a novel adversarial learning framework known as
%
reverse engineering of deception (\textbf{RED}) \cite{darpa2023reverse}. It aims to {infer} the adversary’s information (\textit{e.g.},  the attack objective and adversarial perturbations) from attack instances.
Yet, nearly all the existing {\RED} approaches focused on either estimation/attribution of adversarial perturbations \cite{gong2022reverse,goebel2021attribution,souri2021identification,thaker2022reverse} or recognition of attack classes/types \cite{nicholson2023reverse,wang2023can,maini2021perturbation,zhou2022trace,guo2023scalable}. None of the prior works investigated the feasibility of inferring \textit{{\VM} attributes} from adversarial examples, despite the foundational role of the {\VM} in the attack generation.
Thus, we ask \textbf{(Q)}: 
\begin{center}
    \textit{\textbf{(Q)} 
Can adversarial examples be parsed to reveal {\VM}  information, such as architecture type, kernel size, and 
activation function?}
\end{center}


We refer to the problem encapsulated by question (Q) as 
\textbf{model parsing} of adversarial attacks. For a visual representation of this concept, please refer to \textbf{Fig.\,\ref{fig: overview}} for an illustrative overview.
This work draws inspiration from the concept of model parsing as applied to generative models (\textbf{GM}) \cite{asnani2021reverse}, a process aimed at inferring GM hyperparameters from synthesized photo-realistic images \cite{asnani2021reverse}. Unlike the scenario with GMs, where model attributes are embedded in the generated content, adversarial attacks represent data-specific perturbations formulated through meticulously designed optimizers, not GMs. The `model attributes' subject to extraction from these adversarial instances pertain to the {\VM}, which exhibits a less direct relationship with the perturbed data than the connection between GMs and their synthesized outputs \cite{wang2020cnn,asnani2021reverse, yu2019attributing, frank2020leveraging, guarnera2020deepfake}. Consequently, VM attributes have a subtler influence on the adversarial data, making the task of parsing these attributes inherently more challenging compared to decoding data-independent attributes of GMs.
The proposed model parsing study also has an impact by enabling the inference of  `attack toolchains', in terms of the {\VM} attributes embedded in adversarial attacks. 
This capability aligns with the objectives highlighted in the DARPA RED program, underscoring the strategic importance of understanding and mitigating adversarial tactics \cite{darpa2023reverse}.

\begin{wrapfigure}{r}{0.25\textwidth}
\centerline{
\includegraphics[width=0.24\textwidth]{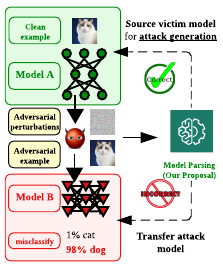}
}
\vspace*{-3.9mm}
\caption{\footnotesize{{
Model parsing in the context of transfer attacks: An effective model parsing system could accurately identify the original {\VM} from which the adversarial attack was generated, as opposed to merely recognizing the target model intended for the transfer attack.
}
}
}
\label{fig: transfer_atk_illustration}
\vspace{-5mm}
\end{wrapfigure}

\noindent
\textbf{A motivational scenario of model parsing through transfer attacks.}
The potential of our model parsing approach can also be demonstrated in the scenario of transfer attacks.
Consider a situation where adversarial examples are crafted using model 
\textit{A} but are employed to compromise model  \textit{B} in a transfer attack setting (refer to \textbf{Fig.\,\ref{fig: transfer_atk_illustration}} for a visual guide). 
Through effective model parsing, it becomes feasible to trace back and identify the original model \textit{A} that served as the source for these adversarial samples, thereby revealing the concealed {\VM} information of the transfer attack.
It's also worth highlight that our investigation is driven by the perspective of a reverse engineer rather than an adversary. Our objective is to deconstruct and understand the origin and characteristics of adversarial examples found in the wild, not to leverage adversarial techniques to extract information from a targeted, opaque model. This distinction underscores our commitment to enhancing security measures rather than undermining them.


\noindent \textbf{Contributions.}
We summarize our contributions below.

\noindent 
$\bullet$ To the best of our knowledge, we for the first time  
 propose and formalize the concept of model parsing to unveil the {VM} attributes from adversarial attacks.

\noindent 
$\bullet$ We approach the model parsing problem of adversarial attacks as a supervised learning task and show that the model parsing network ({\MPN}) could exhibit a surprising amount of generalization   to recognize {\VM} attributes from testing attack instances ({Fig.\,\ref{fig: overview}}). We also peer into the influence of  designing factors (including input data format, backbone network, and evaluation metric) in {\MPN}'s   generalization.

\noindent 
$\bullet$ We make a   comprehensive study on the feasibility and effectiveness of model parsing from adversarial attacks,  including in-distribution generalization ss well as out-of-distribution generalization on unseen attack types and model architectures.  We also demonstrate how the model parsing approach can be used  to  uncover  the true, source  victim model attributes from transfer attacks ({Fig.\,\ref{fig: transfer_atk_illustration}}),
and show a   connection between model parsing   and  attack transferability. 

%% file: secs/related_works.tex
\vspace{-5mm}
\section{Related Work}
 \vspace*{-2mm}

\noindent \textbf{Adversarial attacks and defenses.}
Intensive research efforts have been made for   the design of adversarial attacks and defenses. 
 Adversarial attacks in the digital domain  \cite{goodfellow2014explaining,carlini2017towards,madry2017towards,croce2020reliable,xu2019structured,chen2017ead,xiao2018spatially, liu2018signsgd, chen2017zoo, andriushchenko2020square, brendel2017decision, cheng2019sign, chen2020rays} typically deceive DNNs by integrating  carefully-crafted tiny perturbations  into input data. Adversarial attacks in the physical domain  \cite{eykholt2018robust,li2019adversarial,athalye18b,chen2018shapeshifter,xu2019evading, wang2022survey}   are further developed to fool   victim models under complex physical environmental conditions, which require stronger adversarial perturbations than digital attacks.  
 In this work,  we focus on the commonly-used digital attacks subject to $\ell_p$-norm based perturbation constraints, known as $\ell_p$ attacks.  Based on how an adversary interacts with the {\VM} (victim model), $\ell_p$ attacks also include both perfect-knowledge attacks  (with full access to the {\VM} based on which attacks are generated) and restricted-knowledge attacks  (with access only to the {\VM}'s input and output). 
 The former typically leverages the local gradient information of {\VM} to generate attacks \cite{goodfellow2014explaining, carlini2017towards, madry2017towards}, while the latter takes input-output queries of {\VM} for attack generation \cite{liu2018signsgd, chen2017zoo, andriushchenko2020square,brendel2017decision, cheng2019sign, chen2020rays}. 

 Given the vulnerability of ML models to adversarial attacks, methods to defend against these attacks are another research focus  \cite{madry2017towards,zhang2019theoretically,wong2017provable,salman2020denoised,wong2020fast,carmon2019unlabeled,shafahi2019adversarial,zhang2022revisiting,zhou2022trace,grosse2017statistical,yang2020ml,metzen2017detecting,meng2017magnet,wojcik2020adversarial,liao_defense_2018,xu2019interpreting,srinivasan2021robustifying,shi2021online,yoon2021adversarial,nie2022diffusion, zhang2022distributed}. One line of research is to advance model training methods to acquire adversarially robust models \cite{madry2017towards,zhang2019theoretically,wong2017provable,salman2020denoised,wong2020fast,carmon2019unlabeled,shafahi2019adversarial,zhang2022revisiting, zhang2022distributed}.  Examples include min-max optimization-based adversarial training and its many variants \cite{madry2017towards,zhang2019theoretically,wong2020fast,carmon2019unlabeled,shafahi2019adversarial,zhang2022revisiting}. To make models provably robust,  certified training is also developed   by integrating robustness certificate regularization into model training \cite{boopathy2021fast,raghunathan2018certified,wong2017provable} or  leveraging randomized smoothing
 \cite{salman2020denoised,salman2019provably,cohen2019certified}.
In addition to training robust models, another line of research on adversarial defense is to detect adversarial attacks  by 
exploring and exploiting the differences between adversarial data and benign data
\cite{zhou2022trace,grosse2017statistical,yang2020ml,metzen2017detecting,meng2017magnet,wojcik2020adversarial,liao_defense_2018,xu2019interpreting}. 


\noindent 
\textbf{Reverse engineering of deception ({\RED}).}
 {\RED} has emerged as a new adversarial learning to extract insights into an adversary's strategy, including their identity, objectives, and the specifics of their attack perturbations.
 For example, a few recent works \cite{nicholson2023reverse,wang2023can,maini2021perturbation,zhou2022trace,guo2023scalable} aim to reverse engineer the mechanisms behind attack generation, including the identification of the methods used and the specific hyperparameters (like perturbation radius and step count). 
 In addition, other research efforts, exemplified by works \cite{gong2022reverse,goebel2021attribution,souri2021identification,thaker2022reverse}, have concentrated on estimating or pinpointing the specific adversarial perturbations employed in crafting adversarial imagery.
  This line of research is also related to the area of adversarial purification\cite{srinivasan2021robustifying,shi2021online,yoon2021adversarial,nie2022diffusion}, which aims to mitigate adversarial effects by identifying and eliminating their detrimental impact on model accuracy.
 
 However, none of the prior works investigated
the question of whether attributes of the {\VM} can be reverse-engineered from adversarial attacks.  
The potential to parse {\VM} attributes from adversarial attacks, if realized, could profoundly enhance our comprehension of the underlying threat models. 
Our study draws inspiration from the model parsing concept in GMs (generative models) \cite{asnani2021reverse}, which focuses on inferring GM attributes from their synthesized images. This is based on the premise that GMs embed distinct fingerprints in their outputs, facilitating applications such as DeepFake detection and model attribute inference  \cite{wang2020cnn,asnani2021reverse, yu2019attributing, frank2020leveraging, guarnera2020deepfake}.
%
%
%
Lastly, we stress that  {\RED} diverges from efforts focused on reverse engineering black-box model hyperparameters \cite{oh2019towards, wang2018stealing}, which infer model attributes from a model's prediction logits. Within our model parsing framework, information about the {\VM} is not directly accessible from the adversarial attacks. Our methodology operates without any direct access to the VM, relying solely on adversarial examples gathered from attack generators.

%% file: secs/prob_statement_SLiu.tex
\vspace{-2mm}
\section{Preliminaries and Problem Setups}
\label{sec: problem_formulation}
\vspace{-1mm}


\noindent 
\textbf{Preliminaries: Adversarial attack generation.}
We first introduce different kinds of adversarial attacks  and exhibit their dependence on \textbf{\VM} (\textbf{victim model}), \textit{i.e.}, the ML model from which   attacks are generated. 
Throughout the paper, we will focus on $\ell_{p}$ attacks with $p \in \{ 2, \infty \}$, where the adversary aims to generate imperceptible input perturbations to fool an {image classifier} \cite{goodfellow2014explaining}. Let $\mathbf x$ and $\boldsymbol \theta$ denote a benign image and the parameters of  {\VM}. The \textbf{adversarial attack} (\textit{a.k.a}, adversarial example)   is defined via the linear perturbation  model $\mathbf x^\prime = \mathbf x + \boldsymbol \delta$, where $\boldsymbol \delta = \mathcal A (\mathbf x, \boldsymbol \theta, \epsilon)$ denotes 
 \textbf{adversarial perturbations}, and
$\mathcal{A}$ refers to an attack generation method relying on    $\mathbf x$,   $\btheta$,  and the attack strength $\epsilon$ (\textit{i.e.}, the perturbation radius of $\ell_p$ attacks). 

We focus on {7} attack methods given their different dependencies on      the victim model  ($\btheta$), including
   input gradient-based perfect-knowledge attacks with full access to $\btheta$ ({\FGSM} \cite{goodfellow2014explaining}, {\PGD} \cite{madry2017towards}, {\CW} \cite{carlini2017towards}, and {{\AutoAttack}} or {\AutoAttackShort} \cite{croce2020reliable}) as well as 
  query-based restricted-knowledge attacks 
  ({\ZOSignSGD} \cite{liu2018signsgd},  {\NES} \cite{ilyas2018black}, and {\SquareAttack} or {\SquareAttackShort} \cite{andriushchenko2020square}).
  Among the plethora of $\ell_p$ attack techniques, the methods we have chosen to focus on are characterized by their diverse optimization strategies, loss functions, $\ell_p$ norms, and dependencies on the {\VM}'s parameters ($\btheta$). An overview of these selected methods is presented in \textbf{Table\,\ref{tab: attacks}}.

\begin{table}[htb]
\vspace*{-3mm}
    \centering
     \caption{
      \footnotesize{Summary of focused attack types. Here  GD refers to gradient descent, and PK and RK refer to the perfect-knowledge and restricted-knowledge of the  {\VM}, respectively. 
     }}
      \vspace*{-3mm}
    \label{tab: attacks}
    \resizebox{0.48\textwidth}{!}{
    \begin{tabular}{c | c | c | c | c | c} 
    \toprule[1pt]
    \midrule
    \textbf{Attacks} & \textbf{Generation method} &  \textbf{Loss} & $\ell_{p}$ \textbf{norm} & \textbf{Strength} $\epsilon$ & \textbf{Dependence on $\btheta$}\\
    \midrule
    \FGSM & one-step GD & CE & $\ell_{\infty}$ &\{4, 8, 12, 16\}/255 & PK,  gradient-based \\
    \midrule
    \multirow{2}{*}{\PGD}  & \multirow{2}{*}{multi-step GD} & \multirow{2}{*}{CE}  &  $\ell_\infty$ & \{4, 8, 12, 16\}/255 & \multirow{2}{*}{PK,  gradient-based} \\
     &  &   &   $\ell_2$ &  0.25, 0.5, 0.75, 1  &  \\
    \midrule
    \CW & multi-step GD & CW & $\ell_2$ & 
      \begin{tabular}[c]{@{}c@{}}     
         soft regularization  \\  $c \in \{ 0.1, 1, 10\}$ \end{tabular}
 & PK,  gradient-based \\
    \midrule
   {\AutoAttack}  & \multirow{2}{*}{attack ensemble} & CE /  & $\ell_\infty$   &      
     \{4, 8, 12, 16\}/255  & PK,  gradient-based +   \\
   or {\AutoAttackShort}  &  &  DLR  & $\ell_2$  & 0.25, 0.5, 0.75, 1 & RK, query-based \\
     \midrule
  {\SquareAttack} &  \multirow{2}{*}{random search} & \multirow{2}{*}{CE} & $\ell_\infty$   &  \{4, 8, 12, 16\}/255  & \multirow{2}{*}{RK,  query-based} \\
 or {\SquareAttackShort}    &  &  & $\ell_2$   &  0.25, 0.5, 0.75, 1  &  \\
     \midrule
    \NES & ZOO & CE & $\ell_\infty$ &  \{4, 8, 12, 16\}/255  &  RK, query-based  \\
    \midrule
    \ZOSignSGD & ZOO & CE & $\ell_\infty$ &  \{4, 8, 12, 16\}/255  &  RK, query-based \\
    \midrule
    \bottomrule[1pt]
    \end{tabular}}
     \vspace*{-2mm}
\end{table}

\noindent \ding{70}   {\FGSM} (fast gradient sign  method) \cite{goodfellow2014explaining}: This    attack method   is given by  
  $\bdelta = \mathbf x - \epsilon \times  \mathrm{sign}(\nabla_{\mathbf x} \ell_\mathrm{atk}(\mathbf x; \btheta))$, where $\mathrm{sign}(\cdot)$ is the entry-wise sign operation, and $\nabla_{\mathbf x} \ell_\mathrm{atk}$ is the input gradient of a cross-entropy (CE)-based attack loss $\ell_\mathrm{atk} (\mathbf x; \btheta)$ 

 \noindent \ding{70}    {\PGD} (projected gradient descent)  \cite{madry2017towards}: This extends  {\FGSM} via an iterative algorithm. The $K$-step \textit{{\PGD} $\ell_\infty$ attack} is given by
 $
\bdelta = \bdelta_K$, where 
$\bdelta_k = \mathcal P_{\| \bdelta \|_\infty \leq \epsilon} ( \bdelta_{k-1} - \alpha \times  \mathrm{sign}(\nabla_{\mathbf x} \ell_\mathrm{atk}(\mathbf x; \btheta)) )$ for $k = 1,\ldots, K$, 
 $ \mathcal P_{\| \bdelta \|_\infty \leq \epsilon} $ is the projection operation onto the $\ell_\infty$-norm constraint $\| \bdelta \|_\infty \leq \epsilon$, and $\alpha$ is the attack step size. By replacing the $\ell_\infty$ norm with the $\ell_2$ norm, we   similarly obtain the  \textit{{\PGD} $\ell_2$ attack}  \cite{madry2017towards}. 

\noindent \ding{70} {\CW} (Carlini-Wager)
attack  \cite{carlini2017towards}: Similar to {\PGD}, {\CW} calls   iterative optimization  for attack generation. Yet,  {\CW}  formulates attack generation as an $\ell_p$-norm regularized optimization problem, with the regularization parameter $c =1$  and $p = 2$ by default. Here 
setting the regularization parameter $c=1$ can result in variations in the perturbation strengths ($\epsilon$) across different  {\CIFAR} images. However, the average perturbation strength tends to stabilize around $\epsilon = 0.33$.
Moreover, 
{\CW} adopts a hinge loss to
  ensure the misclassification margin. 

\noindent \ding{70}  {\AutoAttack} (or {\AutoAttackShort}) \cite{croce2020reliable}: This is an ensemble attack  that  uses {\AutoPGD}, an adaptive version of {\PGD}, as  the primary means of attack. The   loss of {\AutoPGD} is   given by the difference of logits ratio (DLR) rather than {CE} or {CW} loss.

\noindent \ding{70} {\ZOSignSGD} \cite{liu2018signsgd} and {\NES} \cite{ilyas2018black}: They are zeroth-order optimization (ZOO)-based restricted-knowledge attacks.
In contrast to perfect-knowledge gradient-based attacks that have full access to the {\VM}'s parameters ($\btheta$), restricted-knowledge attacks interact with the victim model solely through submitting inputs and receiving the corresponding predictions, without direct access to the model's internal structure or gradients.
ZOO   then uses these input-output queries to estimate input gradients and generate adversarial perturbations.  Yet, {\ZOSignSGD} and {\NES}   call different gradient estimators in ZOO \cite{liu2020primer}.


\noindent \ding{70}  {\SquareAttack} (or {\SquareAttackShort}) \cite{andriushchenko2020square}: This attack is built upon random search and thus does not rely on the input gradient of the {\VM}.

It is worth noting that we concentrate on $\ell_\infty$ and $\ell_2$ attacks as our exploration into the potential for model parsing from adversarial examples. \textit{Our aim is not to exhaustively catalog all attack methods but to demonstrate a possibly novel avenue for reverse engineering of {\VM} information carried by adversarial instances.}



\noindent 
\textbf{Model parsing  of adversarial attacks.}
It is clear that adversarial attacks contain the information of  {\VM} ($\btheta$), although the degree of their dependence  varies. Thus, one may wonder if the \textit{attributes} of $\btheta$  can be \textit{inferred} from these  attack instances, \textit{i.e.}, adversarial perturbations/examples. The model attributes of our interest  include  model architectures as well as finer-level knowledge, \textit{e.g.}, activation function type.  We call the resulting problem \textbf{model parsing of adversarial attacks},  as described below.
\begin{tcolorbox}[before skip=2mm, after skip=0.0cm, boxsep=0.0cm, middle=0.0cm, top=0.1cm, bottom=0.1cm]
\vspace*{-0.5mm}
\textbf{(Problem statement)}	Is it possible to infer  {\VM} information from  adversarial attacks? And  what factors will influence such model parsing ability? 
\end{tcolorbox}
\vspace*{2mm}


 To the best of our knowledge, the feasibility  of model parsing for adversarial attacks is an open question. Its challenges stay in two dimensions. \textbf{First}, through the \textbf{model lens}, {\VM} is indirectly coupled with adversarial attacks, \textit{e.g.}, via local gradient information or model queries. Thus,  it remains elusive  what  {\VM} information is fingerprinted  in adversarial attacks and impacts the feasibility of model parsing. \textbf{Second}, through the \textbf{attack lens}, the diversity of adversarial attacks (Table\,\ref{tab: attacks})
  makes a once-for-all model parsing solution extremely difficult. We thus take the first step to investigate the feasibility of model parsing and study what factors may influence its performance. 

\noindent \textbf{Model attributes and setup.} 
We specify {\VM}s as convolutional neural network (CNN)-based image classifiers used by attack generators. 
We consider 5
 CNN  {architecture types} ({\AT}s):  ResNet9, ResNet18, ResNet20, VGG11, and VGG13. Given an {\AT}, CNN models are then configured   by different choices of  {kernel size} ({\KS}),  {activation function} ({\AF}), and {weight  sparsity} ({\WS}). Thus, a valued quadruple
  ({\AT}, {\KS}, {\AF}, {\WS}) 
  yields a specific {\VM} ($\btheta$). 
   \begin{wraptable}{r}{0.3 \textwidth}
 \vspace*{-4mm}
    \centering
     \caption{\footnotesize{Summary of model attributes of interest. Each attribute value corresponds to an attribute class in model parsing. }}
     \label{tab: models}
    \resizebox{0.29\textwidth}{!}{
    \begin{tabular}{c | c | c} 
    \toprule[1pt]
    \midrule
    \textbf{Model attributes} & \textbf{Code} &  \textbf{Classes per attribute} \\
    \midrule
    \multirow{2}{*}{Architecture type} & \multirow{2}{*}{\AT}  & ResNet9, ResNet18 \\
     &   & ResNet20, VGG11, VGG13  \\
    \midrule
    Kernel size & \KS  & 3, 5, 7 \\
    \midrule
    Activation function & \AF & ReLU, tanh, ELU \\
    \midrule
    Weight sparsity & \WS & 0\%, 37.5\%, 62.5\% \\
    \midrule
    \bottomrule[1pt]
    \end{tabular}}
     \vspace*{-5mm}
\end{wraptable}
  Although more attributes could be considered, we focus on {\KS} and {\AF}  since they are the two fundamental building components of  CNNs. 
Besides, 
 we choose {\WS} as another model attribute since it relates to sparse models achieved by pruning (\textit{i.e.}, removing redundant model weights) \cite{han2015learning,frankle2018lottery}. 
 
  \textbf{Table\,\ref{tab: models}} summarizes the model attributes and their values when  specifying    {\VM} instances. 
Given a {\VM} specification,  adversarial attacks are generated following Table\,\ref{tab: attacks}.  

%% file: secs/methods_SLiu.tex
\vspace{-2mm}
\section{Methods}
\label{sec: Methods}
\vspace{-1mm}

In this section,
we approach the model parsing problem as a supervised learning task applied over the dataset of adversarial attacks. 
We will show that the learned model could  exhibit a surprising amount of generalization   on test-time adversarial data. We will also  show     data-model factors  that may  influence such generalization.


\noindent 
\textbf{Model parsing network and training.} We propose a parametric model, termed model parsing network (\textbf{\MPN}),  which takes adversarial attacks as input and predicts the model attribute values (\textit{i.e.}, `classes' in Table\,\ref{tab: models}).  
It is worth noting that the proposed {\MPN} operates solely on adversarial examples, possessing no prior information about the victim model, highlighting its capacity to unveil the secretes of {\VM} embedded in adversarial examples. 
Despite the simplicity of supervised learning,  the construction of {\MPN} is non-trivial considering the factors such as  the input data format, the choice of an appropriate backbone network, and the determination of suitable evaluation metrics.

First, we create a dataset  by collecting adversarial examples against {\VM}s. 
Since adversarial attacks are proposed for evading model predictions after training, we choose the test set of an ordinary image dataset (\textit{e.g.}, {\CIFAR})  to generate adversarial data, where an 80/20 training/test split is used for {\MPN} training and evaluation. 
Following notations in Sec.\,\ref{sec: problem_formulation}, the training set  of  {\MPN}   is  denoted by  $\mathcal D_\mathrm{tr} = \{ (\mathbf z( \mathcal A, \mathbf x, \btheta ), y(\btheta)) \, | \, \mathbf x \in \mathcal I_\mathrm{tr},  \btheta \in  \Theta \}$, where 
$\mathbf z$ denotes attack instances (\textit{e.g.}, adversarial perturbations $\boldsymbol \delta$ or adversarial example $\mathbf x^\prime$) that relies on the attack method $\mathcal A$, the original image sample $\mathbf x$, and the {\VM} $\btheta$, and $y(\btheta)$ denotes the true model attribute label of $\btheta$ associated with $\mathbf z$. To differentiate with  the testing data of {\MPN}, we denote by  
$\mathcal I_\mathrm{tr}$   the set of original images used for training {\MPN}. We also denote by $\Theta$   the set of {\VM}s used for generating adversarial examples. 
For  simplicity, we denote the training set of {\MPN} as 
$\mathcal D_\mathrm{tr} = \{ (\mathbf z, y) \}$  to omit the   dependence on other factors.

\begin{wrapfigure}{r}{0.3 \textwidth}
\vspace{-5mm}
\centerline{
\includegraphics[width=0.29\textwidth]{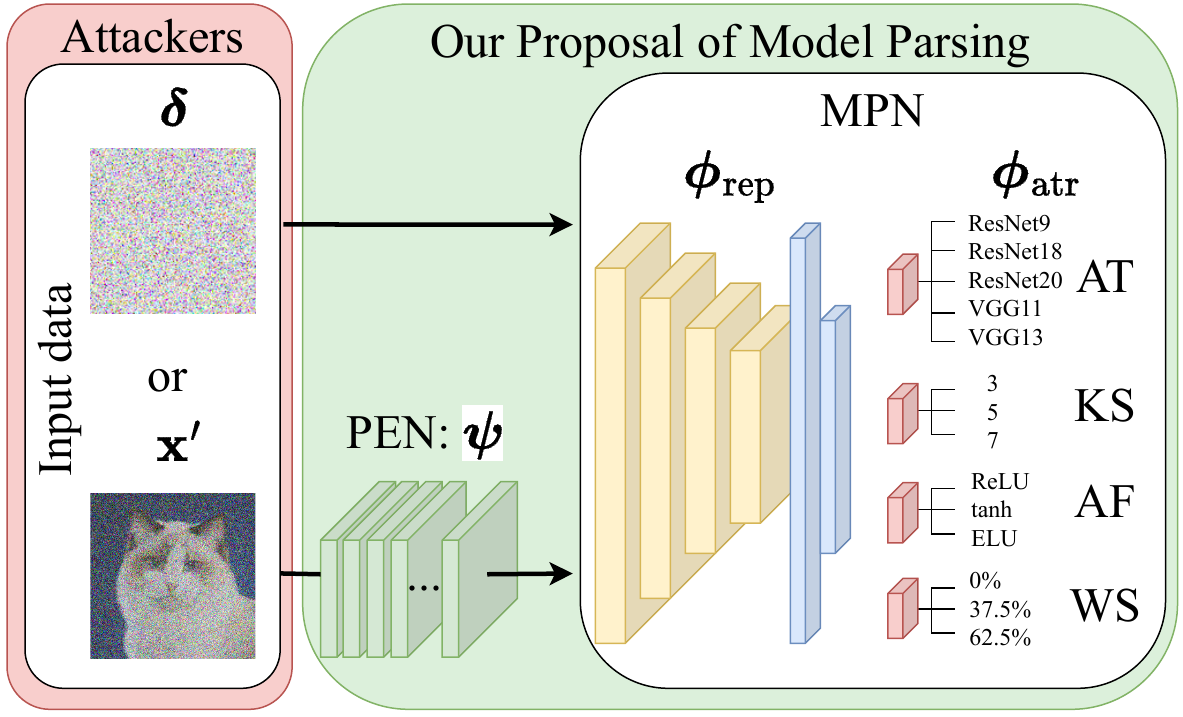}}
\vspace*{-2mm}
\caption{\footnotesize{Model parsing via supervised learning. 
Adversarial examples or perturbations, crafted by attackers, serve as the input of {\MPN}, which aims to decode {\VM} attributes from adversarial inputs. The {\PEN} (perturbation estimation network), introduced subsequently, acts as a preprocessing step, converting adversarial examples into inputs resembling perturbations.
}}
\label{fig: overview_classifier}
\vspace{-5mm}
\end{wrapfigure}
Next, we study the construction of   {\MPN} (parameterized by $\boldsymbol \phi$). 
First, we manage to examine the \textit{feasibility} of model parsing even forcing the  \textit{simplicity} of attribution network. Second, we manage to avoid the \textit{model attribute bias} of {$\boldsymbol \phi$} when inferring  {\VM}  attributes. 
Therefore, we specify     {\MPN} by two  simple networks: (1) multilayer perceptron ({\MLP}) containing 2 hidden layers with 128 hidden units (0.41M parameters) \cite{lecun2015deep}, and (2) a simple 4-layer CNN   ({\Conv}) with 64 output channels for each layer, followed by one fully-connected layers with 128 hidden units and the attribution prediction head (0.15M parameters) \cite{vinyals2016matching}.
We found that the model parsing accuracy using {\Conv} typically outperforms that of {\MLP}. 
Thus,  {\Conv}   is designated as the default architecture for our {\MPN}.

Given the datamodel setup, we next tackle the recognition problem of  {\VM}'s attributes ({\AT}, {\KS}, {\AF}, {\WS}) via  a multi-head multi-class classifier. We   dissect    {\MPN}    into two parts  $\boldsymbol \phi = [{\boldsymbol \phi}_\text{rep}, \boldsymbol \phi_\text{atr}]$, where ${\boldsymbol \phi}_\text{rep}$ is   for  data \underline{rep}resentation acquisition,  and ${\boldsymbol \phi}_\text{attr}$ corresponds to the \underline{attr}ibute-specific prediction head (\textit{i.e.}, the last fully-connected layer in our design). Eventually, four prediction heads $\{ \boldsymbol \phi_\text{atr}^{(i)}\}_{i=1}^4$ will share ${\boldsymbol \phi}_\text{rep}$ for model attribute recognition; see \textbf{Fig.\,\ref{fig: overview_classifier}} for a schematic overview of our proposal. The {\MPN} training problem is then cast as 

\vspace*{-5mm}
{\small \begin{align}
     \hspace*{-2mm}  \begin{array}{l}
 \displaystyle \minimize_{{\boldsymbol \phi}_\text{rep}, \{ \boldsymbol \phi_\text{atr}^{(i)}\}_{i=1}^4}         \mathbb E_{(\mathbf z, y) \in \mathcal D_\mathrm{tr}} \sum_{i=1}^4 [ \ell_{\mathrm{CE}} ( h (  \mathbf z; {\boldsymbol \phi}_\text{rep}, \boldsymbol \phi_\text{atr}^{(i)}   ), y_i ) ], 
    \end{array}
  \hspace*{-2mm} 
  \label{eq: train_supervision}
\end{align}}%
where $h (  \mathbf z; {\boldsymbol \phi}_\text{rep} , \boldsymbol \phi_\text{atr}^{(i)}   )$ denotes the {\MPN} prediction at input example $ \mathbf z$ using the predictive model consisting of ${\boldsymbol \phi}_\text{rep} $ and  $ \boldsymbol \phi_\text{atr}^{(i)}$ for the $i$th attribute classification, $y_i$ is the ground-truth label of the $i$th attribute  associated with the input data $\mathbf z$, and
$\ell_{\mathrm{CE}}$ is the cross-entropy (CE) loss   characterizing the   error between the prediction and the true label.

\noindent \textbf{Evaluation methods.}
Similar to    training, we denote by $\mathcal D_\mathrm{test} = \{ (\mathbf z( \mathcal A, \mathbf x, \btheta ), y(\btheta)) \, | \, \mathbf x \in \mathcal I_\mathrm{test},  \btheta \in  \Theta \}$ the test attack set  for evaluating the performance of    {\MPN}. Here the set of benign images $\mathcal I_\mathrm{test}$ is different from $\mathcal I_\mathrm{tr}$, thus   adversarial attacks in $\mathcal D_\mathrm{test}$ are new to $\mathcal D_\mathrm{tr}$.
To mimic the standard evaluation pipeline of supervised learning, we propose the following evaluation metrics.

\textit{(1) In-distribution generalization}:  The {\MPN} testing dataset $\mathcal D_\mathrm{test}$ follows the attack methods ($\mathcal A$) and the {\VM} specifications ($\Theta$)  \textit{same as} $\mathcal D_\mathrm{tr}$ but corresponding to different benign images (\textit{i.e.}, $\mathcal I_\mathrm{test} \neq \mathcal I_\mathrm{tr}$). 
The purpose of such an in-distribution evaluation is to examine if the trained {\MPN} can infer model attributes encoded in new attack data given existing attack methods.

\textit{(2) Out-of-distribution (OOD) generalization}: 
In addition to  new test-time images, 
there exist \textit{attack/model distribution shifts} in $\mathcal D_\mathrm{test}$
due to using  \textit{new} attack methods 
or model architectures, leading to  \textit{unseen} attack methods ($\mathcal A$) and    victim models ($\Theta$) different from the settings   in  $\mathcal D_\mathrm{tr}$.

 Unless specified otherwise,   the generalization of {\MPN} stands for the \textit{in-distribution generalization}. Yet, both in-distribution and OOD generalization capabilities will be empirically assessed.



\begin{wrapfigure}{r}{0.26\textwidth}
\vspace*{-9mm}
\centerline{
\includegraphics[width=0.24\textwidth]{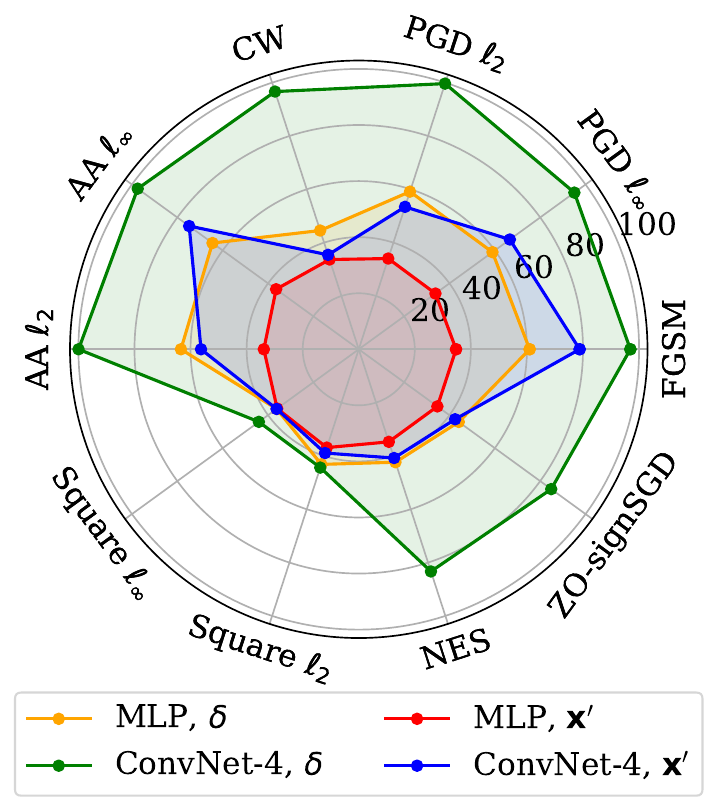}
}
\vspace*{-3mm}
\caption{\footnotesize{
The {\VM} attribute classification accuracy  of {\MPN} under different input formats (adversarial perturbations $\bdelta$ vs. examples $\mathbf x^\prime$) and parsing networks ({\Conv} vs. {\MLP}). The accuracy is measured in the context of in-distribution generalization. 
The attack data is generated from attack methods given in  Table\,\ref{tab: attacks},  with $\ell_\infty$ attack strength $\epsilon = 8/255$ and $\ell_2$ attack strength $\epsilon = 0.5$ on {\CIFAR}. 
}}
\vspace*{-6mm}
\label{fig: results_MPN_models_inputs_1}
\end{wrapfigure}
\noindent  \textbf{Perturbations or adversarial examples? The input data format matters for {\MPN}.}
Recall from Sec.\,\ref{sec: problem_formulation} that 
an adversarial example, given by the linear   model $\mathbf x^\prime = \mathbf x + \bdelta$, relates to $\btheta$ through $\bdelta$. 
Thus, it could be better for {\MPN} to adopt     \textit{adversarial perturbations} ($\boldsymbol \delta$) as the attack data feature ($\mathbf z$),  rather than the indirect adversarial example $\mathbf x^\prime$.
\textbf{Fig.\,\ref{fig: results_MPN_models_inputs_1}} empirically justifies our hypothesis by comparing the generalization  of  {\MPN} trained on adversarial perturbations with that   on adversarial examples under two    model specifications of {\MPN},  {\MLP} and {\Conv}.
We present the performance of {\MPN} trained and tested on different   attack types. As we can see, the use of adversarial perturbations ($\boldsymbol \delta$)    consistently improves the classification accuracy of {\VM} attributes, compared to the use of adversarial examples ($\mathbf x^\prime$). In addition,   {\Conv} outperforms {\MLP} with a substantial margin.

Although Fig.\,\ref{fig: results_MPN_models_inputs_1}  shows  the promise of the generalization ability of {\MPN} when trained and tested on adversarial perturbations, it may raise another practical question of how to obtain adversarial perturbations from adversarial examples if the latter is the only  attack source  accessible to   {\MPN}. 
To overcome this difficulty, we propose a \textit{perturbation estimator network} (\textbf{\PEN})  that can be jointly learned with {\MPN}.
Once {\PEN} is prepended to the {\MPN} model, the resulting end-to-end pipeline can achieve model parsing using  adversarial examples as inputs (see the lower pipeline in Fig.\,\ref{fig: overview_classifier}). 
We use a denoising network, DnCNN \cite{zhang2017beyond}, to model {\PEN} with parameters $\boldsymbol \psi $. 
{\PEN} obtains perturbation estimates by minimizing the denoising objective using the true adversarial perturbations as   supervision. Extended from \eqref{eq: train_supervision}, we have

\vspace*{-5mm}
{\small \begin{align}
   \hspace*{-5mm} \begin{array}{ll}
   \displaystyle \minimize_{\bpsi, {\boldsymbol \phi}_\text{rep}, \{ \boldsymbol \phi_\text{atr}^{(i)}\}_{i=1}^4}     & \beta \mathbb E_{(\mathbf x, \mathbf x^\prime) \in \mathcal D_\mathrm{tr}}  [ \ell_{\mathrm{MAE}}(g_{\bpsi}(\mathbf x^\prime), \mathbf x^{\prime} - \mathbf x) ] 
   \\
        & \hspace*{-20mm}+ 
        \mathbb E_{(\mathbf x^\prime,  y) \in \mathcal D_\mathrm{tr}} \sum_{i=1}^4 [ \ell_{\mathrm{CE}} ( h ( g_{\bpsi}( \mathbf x^\prime) ; {\boldsymbol \phi}_\text{rep}, \boldsymbol \phi_\text{atr}^{(i)}   ), y_i ) ],
   \end{array}
   \hspace*{-5mm}
    \label{eq: pen_formulation}
\end{align}}%
where $g_{\bpsi}(\mathbf{x}^\prime)$ is output of {\PEN} given  $\mathbf x^{\prime}$ as input, $\ell_{\mathrm{MAE}}$ is the mean-absolute-error (MAE) loss characterizing the perturbation estimation error, and $\beta > 0$ is a regularization parameter. Compared with \eqref{eq: train_supervision}, {\MPN}   is integrated with the perturbation estimation $g_{\bpsi}( \mathbf x^\prime)$ for {\VM} attribute classification.

%% file: secs/experiments_SLiu.tex
\vspace{-2mm}
\section{Experiments}
\label{sec: exp}
\vspace{-1mm}


\subsection{Experiment setup and implementation}
\label{sec: exp_setup}

\noindent \textbf{Dataset curation.}
We use standard image classification datasets ({\CIFAR}, {\CIFARM}, and {\TImageNet})  to train {\VM}s, from which attacks are generated.  We refer readers to Appendix\,\ref{apx_sec:victim_model_train} for details on {\VM} training and evaluation, as well as different attack setups. These {\VM} instances are then leveraged to create the training and evaluation datasets of {\MPN}, as described in Sec.\,\ref{sec: Methods}. The attack types and victim model configurations have been summarized in  Table\,\ref{tab: attacks} and \ref{tab: models}. Eventually, we collect a dataset consisting of adversarial attacks across $7$ attack types generated from $135$ {\VM}s (configured by $5$ architecture types, $3$  kernel size setups, $3$ activation function types, and $3$ weight sparsity levels).


\noindent
\textbf{{\MPN} training and evaluation.}
To solve problem \eqref{eq: train_supervision}, we train the {\MPN} model using the SGD (stochastic gradient descent) optimizer with cosine annealing learning rate schedule and an initial learning rate of $0.1$. The training epoch number and the batch sizes are given by $100$ and  $256$, respectively. 
To solve problem  \eqref{eq: pen_formulation},  we first train {\MPN} according to \eqref{eq: train_supervision}, and then fine-tune a pre-trained DnCNN model \cite{gong2022reverse} (taking only the denoising objective into consideration)  for 20 epochs.  Starting from these initial models, we jointly optimize {\MPN} and {\PEN}  by minimizing problem  \eqref{eq: pen_formulation} with   $\beta = 1$ over 50 epochs.
To evaluate the effectiveness of {\MPN}, we consider both   in-distribution   and OOD generalization assessment. The generalization performance is measured by testing accuracy    averaged over  attribute-wise predictions, namely, $\sum_i (N_i \mathrm{TA} (i) ) / \sum_i N_i$, where $N_i$ is the number of classes of the model attribute $i$, and $ \mathrm{TA} (i)$ is the testing accuracy of  the classifier associated with the attribute $i$. 


\subsection{Results and insights}
\vspace*{-1mm}

\noindent
\textbf{In-distribution generalization of {\MPN} is  achievable.}
\textbf{Table\,\ref{tab: in_distribution}} presents the in-distribution generalization performance of {\MPN} trained using different input data formats (\textit{i.e.}, adversarial examples $\mathbf x^\prime$, {\PEN}-estimated adversarial perturbations $\bdelta_\text{\PEN}$, and true adversarial perturbations $\bdelta$) given each attack type in Table\,\ref{tab: attacks}. Here the choice of {\AT} (architecture type)
is fixed to ResNet9, but  adversarial attacks on  {\CIFAR} are generated  from {\VM}s  configured by   different values of {\KS}, {\AF}, and {\WS} (see Table\,\ref{tab: models}).
As we can see, the generalization of {\MPN}   varies  against the attack type even if model parsing is conducted from the ideal adversarial perturbations   ($\bdelta$). We also note that model parsing from perfect-knowledge adversarial attacks (\textit{i.e.}, {\FGSM}, {\PGD}, and {\AutoAttackShort}) is easier than that from restricted-knowledge attacks (\textit{i.e.}, {\ZOSignSGD}, {\NES}, and {\SquareAttackShort}). For example, the worst-case performance of {\MPN} is achieved when  training/testing on {\SquareAttackShort}  attacks. This is not surprising, since {\SquareAttackShort} is based on random search and has the least dependence on {\VM} attributes. 
In addition, we find that {\MPN} using estimated perturbations  ($\bdelta_\mathrm{PEN}$) substantially outperforms the one   trained   on adversarial examples ($\mathbf x^\prime$). This justifies the effectiveness of {\PEN} solution for {\MPN}. 

\begin{table}[t]
\vspace{0mm}
    \centering
    \caption{\footnotesize{The in-distribution testing accuracy (\%) of {\MPN}  trained  using different input data formats (adversarial examples $\mathbf x^\prime$, {\PEN}-estimated adversarial perturbations $\bdelta_\text{\PEN}$, and true adversarial perturbations $\bdelta$) across   different attack types on {\CIFAR}, with   $\ell_\infty$ attack strength $\epsilon = 8/255$,  $\ell_2$ attack strength $\epsilon = 0.5$, and {\CW} attack strength $c = 1$.  
    }}
    \label{tab: in_distribution}
    \vspace{-3mm}
       \resizebox{0.48\textwidth}{!}{
    \begin{tabular}{c|c|c|c|c|c|c|c|c|c|c}
\toprule[1pt] 
\midrule
\diagbox{\textbf{Input data}}{\textbf{Attack type}} & {\FGSM} & \begin{tabular}[c]{@{}c@{}}     
      {\PGD} 
      $\ell_\infty$
      \end{tabular}
      & 
      \begin{tabular}[c]{@{}c@{}}     
      {\PGD} 
      $\ell_2$
      \end{tabular} 
      & {\CW} & 
      \begin{tabular}[c]{@{}c@{}}     
      {\AutoAttackShort} 
      $\ell_\infty$
      \end{tabular} & 
    \begin{tabular}[c]{@{}c@{}}     
      {\AutoAttackShort} 
      $\ell_2$
      \end{tabular}
      & \begin{tabular}[c]{@{}c@{}}     
      {\SquareAttackShort} 
      $\ell_\infty$
      \end{tabular} & 
\begin{tabular}[c]{@{}c@{}}     
      {\SquareAttackShort} 
      $\ell_2$
      \end{tabular}
      & {\NES} &  \begin{tabular}[c]{@{}c@{}}     
      {\texttt{ZO}-} 
      {\texttt{signSGD}}
      \end{tabular} 
\\ \midrule
$\mathbf{x}^\prime$         &78.80  &66.62  &53.42  &35.42  &74.78  &56.26  &38.92  &36.21  &40.80  &42.48 \\ \midrule
$\bdelta_{\PEN}$         &94.15  &83.20  &82.58  &64.46  &91.09  &86.89  &44.14  &42.30  &58.85  &61.20 \\ \midrule
$\bdelta$         &96.89  &95.07  &99.64  &96.66  &97.48  &99.95  &44.37  &44.05  &83.33  &84.87 \\ \midrule\bottomrule[1pt]
\end{tabular}}
\vspace*{-5mm}
\end{table}

\begin{table*}[th]
\centering
\caption{\footnotesize{
In-distribution generalization performance (testing accuracy, \%) of {\MPN} given different choices of {\VM}s and datasets, attack types/strengths, and {\MPN} input data formats ($\mathbf x^\prime$, $\bdelta_\mathrm{PEN}$, and $\bdelta$). 
}}
    \label{tab: in_distribution_arch_selected}
\vspace{-3mm}
\resizebox{\textwidth}{!}{
\begin{tabular}{l|l|ccc|ccc|ccc|ccc|ccc|ccc|ccc}
\toprule[1pt]\midrule
\multirow{4}{*}{\begin{tabular}[c]{@{}c@{}}\textbf{Attack}\\\textbf{ type}\end{tabular}} & 
\multirow{4}{*}{\begin{tabular}[c]{@{}c@{}} \textbf{ Attack}\\\textbf{  strength} \end{tabular}}
& \multicolumn{18}{c}{\textbf{Dataset and victim model}} \\
 & &\multicolumn{3}{c|}{\begin{tabular}[c]{@{}c@{}}{\CIFAR}\\ ResNet9\end{tabular}}&\multicolumn{3}{c|}{\begin{tabular}[c]{@{}c@{}}{\CIFAR}\\ ResNet18\end{tabular}} & \multicolumn{3}{c|}{\begin{tabular}[c]{@{}c@{}}{\CIFAR}\\ ResNet20\end{tabular}} & \multicolumn{3}{c|}{\begin{tabular}[c]{@{}c@{}}{\CIFAR}\\ VGG11\end{tabular}} & \multicolumn{3}{c|}{\begin{tabular}[c]{@{}c@{}}{\CIFAR}\\ VGG13\end{tabular}} & \multicolumn{3}{c|}{\begin{tabular}[c]{@{}c@{}}{\CIFARM}\\ ResNet9\end{tabular}} & \multicolumn{3}{c}{\begin{tabular}[c]{@{}c@{}}{\TImageNet}\\ ResNet18\end{tabular}} \\ 
 & &\multicolumn{1}{c|}{$\mathbf {x}^\prime$} & \multicolumn{1}{c|}{$\bdelta_\mathrm{PEN}$} & \multicolumn{1}{c|}{$\bdelta$}  & \multicolumn{1}{c|}{$\mathbf {x}^\prime$} & \multicolumn{1}{c|}{$\bdelta_\mathrm{PEN}$} & \multicolumn{1}{c|}{$\bdelta$} & \multicolumn{1}{c|}{$\mathbf {x}^\prime$} & \multicolumn{1}{c|}{$\bdelta_\mathrm{PEN}$} & \multicolumn{1}{c|}{$\bdelta$} & \multicolumn{1}{c|}{$\mathbf {x}^\prime$} & \multicolumn{1}{c|}{$\bdelta_\mathrm{PEN}$} & \multicolumn{1}{c|}{$\bdelta$} & \multicolumn{1}{c|}{$\mathbf {x}^\prime$} & \multicolumn{1}{c|}{$\bdelta_\mathrm{PEN}$} & \multicolumn{1}{c|}{$\bdelta$} & \multicolumn{1}{c|}{$\mathbf {x}^\prime$} & \multicolumn{1}{c|}{$\bdelta_\mathrm{PEN}$} & \multicolumn{1}{c|}{$\bdelta$} & \multicolumn{1}{c|}{$\mathbf {x}^\prime$} & \multicolumn{1}{c|}{$\bdelta_\mathrm{PEN}$} & \multicolumn{1}{c}{$\bdelta$} \\ \midrule
\multirow{4}{*}{\begin{tabular}[c]{@{}c@{}}{\FGSM}\end{tabular}} 
&
$\epsilon = 4/255$ 
        &60.13  &85.25  &96.82  &60.00  &86.92  &97.66  &62.41  &88.91  &97.64  &47.42  &73.40  &91.75  &66.28  &90.02  &98.57  &57.99  &82.22  &94.86  &37.23  &84.27  &97.04 \\
& $\epsilon = 8/255$ 
        &78.80  &94.15  &96.89  &80.44  &95.49  &97.61  &82.29  &95.90  &97.72  &63.13  &86.76  &92.41  &84.92  &96.91  &98.66  &75.58  &91.65  &94.96  &70.29  &91.17  &97.05 \\
& $\epsilon = 12/255$ 
        &86.49  &95.96  &96.94  &88.03  &96.89  &97.68  &88.71  &97.13  &97.81  &73.71  &90.19  &92.66  &91.21  &98.10  &98.71  &82.27  &94.01  &95.55  &76.00  &93.45  &97.02 \\
& $\epsilon = 16/255$ 
        &90.16  &96.43  &96.94  &91.71  &97.34  &97.68  &91.84  &97.47  &97.79  &79.51  &91.28  &92.60  &94.22  &98.44  &98.73  &86.50  &94.04  &94.74  &79.63  &94.35  &96.87 \\
\midrule
\multirow{4}{*}{{\PGD} $\ell_\infty$} 
&$\epsilon = 4/255$ 
        &50.54  &76.43  &96.02  &56.94  &79.45  &96.96  &55.01  &80.05  &97.49  &39.33  &66.38  &91.84  &57.12  &81.18  &98.29  &42.27  &72.62  &92.65  &35.48  &76.56  &97.18\\
&$\epsilon = 8/255$ 
        &66.62  &83.20  &95.07  &73.29  &87.29  &95.38  &67.49  &86.19  &96.18  &56.62  &81.14  &92.78  &69.16  &88.46  &97.22  &59.71  &79.55  &90.43  &61.85  &82.90  &96.05 \\
&$\epsilon = 12/255$ 
        &76.65  &89.73  &94.91  &81.73  &91.67  &95.55  &76.41  &90.16  &95.67  &70.56  &88.92  &94.13  &78.67  &92.93  &97.26  &70.86  &85.31  &91.28  &73.82  &88.80  &96.38 \\
&$\epsilon = 16/255$
        &75.58  &86.95  &91.28  &82.46  &90.19  &93.19  &76.58  &87.79  &92.50  &72.13  &87.23  &91.85  &78.28  &90.20  &94.66  &71.29  &82.35  &86.84  &73.19  &85.02  &93.54 \\
\midrule
\multirow{4}{*}{{\PGD} $\ell_2$} 
&$\epsilon = 0.25$
        &36.75  &62.20  &99.66  &46.35  &70.17  &99.74  &48.24  &77.22  &99.75  &36.47  &45.17  &98.52  &35.81  &70.62  &99.85  &35.92  &61.91  &99.29  &35.55  &35.68  &99.68 \\
&$\epsilon = 0.5$
        &53.42  &82.58  &99.64  &60.89  &84.70  &99.56  &61.62  &89.11  &99.61  &41.56  &66.58  &98.68  &57.83  &87.64  &99.83  &48.89  &79.26  &99.01  &35.52  &54.56  &99.71\\
&$\epsilon = 0.75$
        &62.66  &89.04  &99.48  &71.01  &89.89  &99.22  &70.76  &92.06  &99.36  &47.02  &78.12  &98.52  &72.76  &92.32  &99.74  &59.19  &85.14  &98.61  &35.56  &81.33  &99.71 \\
&$\epsilon = 1$
        &71.65  &91.73  &99.26  &77.09  &92.09  &98.94  &76.84  &92.82  &98.96  &54.20  &84.30  &98.41  &79.93  &93.96  &99.57  &66.97  &87.63  &97.89  &43.48  &88.81  &99.64  \\
\midrule
\multirow{3}{*}{\CW} 
&$c = 0.1$
        &33.77  &55.60  &96.71  &47.77  &63.26  &96.11  &33.56  &63.11  &94.10  &33.73  &48.90  &94.37  &33.68  &65.48  &96.95  &34.41  &46.47  &92.55  &35.96  &35.77  &95.52  \\
&$c = 1$
        &35.42  &64.46  &96.66  &45.75  &65.25  &97.45  &33.74  &62.71  &97.08  &33.89  &55.61  &91.29  &36.12  &68.66  &98.58  &34.25  &55.18  &93.25  &35.54  &35.29  &89.35 \\
&$c = 10$
        &36.38  &64.45  &96.64  &45.83  &65.32  &97.41  &33.83  &63.52  &97.11  &38.29  &56.83  &91.33  &38.51  &68.28  &98.62  &34.25  &55.89  &93.18  &35.45  &53.18  &94.20  \\
\midrule\bottomrule[1pt]
\end{tabular}}
\vspace{-6mm}
\end{table*}

\begin{figure}
\centerline{
\hspace*{2mm}\includegraphics[width=0.35\textwidth,height=!]{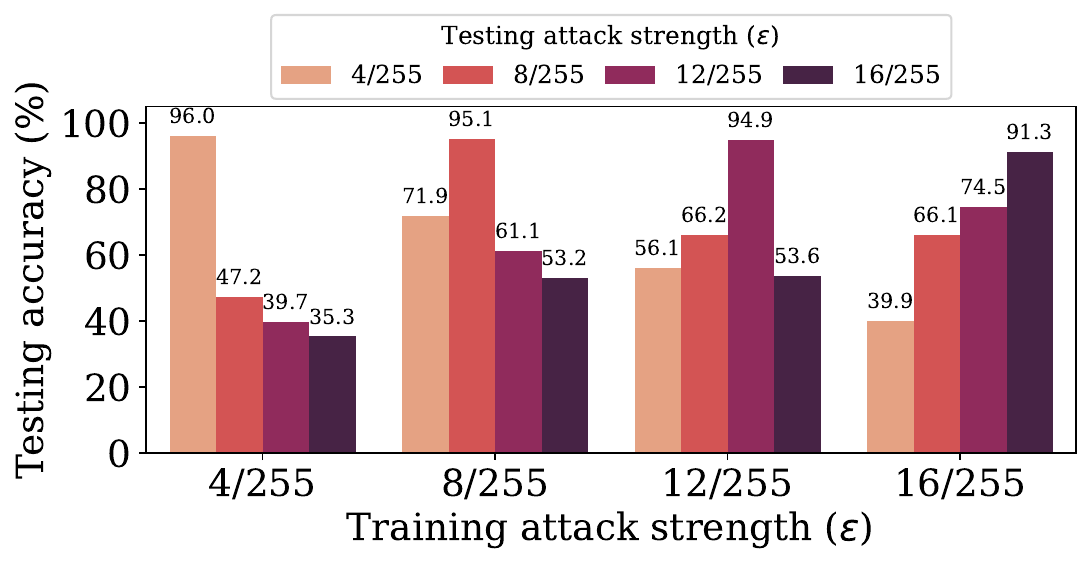}
}
\vspace*{-1mm}
\caption{\footnotesize{
{Testing accuracies  (\%) of {\MPN} when trained on   adversarial perturbations generated by {\PGD} $\ell_\infty$ using different attack strengths ($\epsilon$) and evaluated using different attack strengths as well. Other setups are consistent with  in Table\,\ref{tab: in_distribution}.}
}}
\label{fig: mismatch_attack_strength_selected}
 \vspace*{-7mm}
\end{figure}
Extended from Table\,\ref{tab: in_distribution}, \textbf{Fig.\,\ref{fig: mismatch_attack_strength_selected}} shows the generalization performance of {\MPN} when evaluated using  attack data with different attack strengths. We observe that  in-distribution generalization {(corresponding to the same attack strength for the train-time and test-time attacks)}
is easier to achieve than OOD generalization (different attack strengths at  test time and train time).
Another observation is that a smaller gap between the train-time attack strength and the test-time strength leads to better generalization performance.


Extended from Table\,\ref{tab: in_distribution} and Fig.\,\ref{fig: mismatch_attack_strength_selected} that focused on model parsing of adversarial attacks by fixing the {\VM} architecture to ResNet9 on {\CIFAR}, 
\textbf{Table\,\ref{tab: in_distribution_arch_selected}}     shows the generalization of {\MPN} under diverse setups of victim model architectures and datasets.  The insights into model parsing   are consistent with Table\,\ref{tab: in_distribution}: (1) The use of true adversarial perturbations ($\bdelta$) and {\PEN}-estimated perturbations ($\bdelta_\mathrm{PEN}$) can yield higher  model parsing accuracy; (2) Inferring model attributes from perfect-knowledge, gradient-based adversarial perturbations  is easier, as supported by its over 90\% testing accuracy; And (3) the model parsing accuracy gets better if adversarial attacks have a higher attack strength ($\epsilon$).

\begin{figure}
\centerline{
\includegraphics[width=0.45\textwidth]{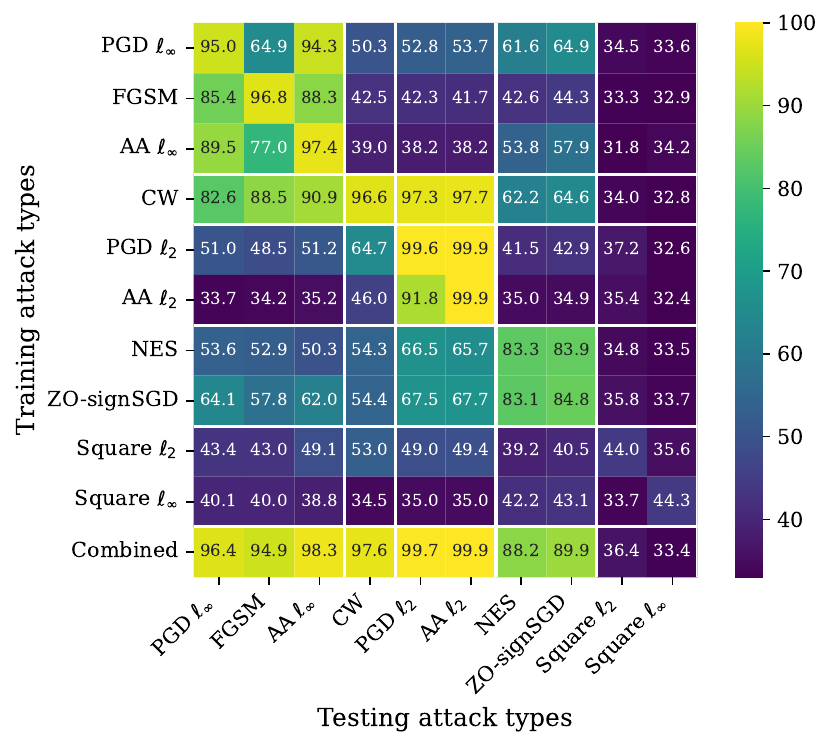}
}
\vspace{-3mm}
\caption{\footnotesize{
Model parsing accuracy (\%) of {\MPN} when trained on a row-specific attack type but evaluated on a column-specific attack type. The attack generation and data-model setups are consistent with Table\,\ref{tab: in_distribution}. {\MPN} takes  adversarial perturbations as input. 
`Combined' represents {\MPN} trained on multiple attack types: {\PGD} $\ell_\infty$, {\PGD} $\ell_2$, {\CW}, and {\ZOSignSGD}. 
}}
\label{fig: id_vs_ood_revised}
\vspace{-7mm}
\end{figure}
\noindent
\textbf{OOD generalization of {\MPN} is difficult vs. unseen attack types at test time.}
In \textbf{Fig.\,\ref{fig: id_vs_ood_revised}}, we present the model parsing accuracy of {\MPN} when trained under one attack type (\textit{e.g.}, {\PGD} $\ell_\infty$ attack at row $1$) but tested under another attack type (\textit{e.g.}, {\FGSM} attack at column $2$) 
on {\CIFAR}.   The diagonal entries of the   matrix correspond to the in-distribution generalization of {\MPN} given the attack type, while the off-diagonal entries denote OOD generalization when test-time attack types are different from train-time ones.

\textbf{First}, we find that {\MPN} generalizes better across attack types when they share similarities, leading to the following \textit{generalization communities}: 
 $\ell_\infty$ attacks ({\PGD} $\ell_\infty$, {\FGSM}, and {\AutoAttackShort} $\ell_\infty$), $\ell_2$ attacks ({\CW}, {\PGD} $\ell_2$, or {\AutoAttackShort} $\ell_2$), and ZOO-based restricted-knowledge attacks ({\NES} and {\ZOSignSGD}).
\textbf{Second}, {\SquareAttackShort} attacks are difficult to learn and generalize, as evidenced by the low test accuracies in the last two rows and   the last two columns. This is also consistent with Table\,\ref{tab: in_distribution}.  
\textbf{Third}, given the existence of generalization communities, we then     combine diverse attack types  (including {\PGD} $\ell_\infty$, {\PGD} $\ell_2$, {\CW}, and {\ZOSignSGD}) into an augmented   {\MPN} training set and investigate if  such a data augmentation  can boost the OOD generalization of   {\MPN}.  The results are summarized in the  \textbf{`combined' row  of Fig.\,\ref{fig: id_vs_ood_revised}}. As we  expect,   the use of combined attack types indeed makes {\MPN} generalize better across all attack types except for the random search-based {\SquareAttackShort} attack.

In Fig.\,\ref{fig: denoise_ood} of Appendix\,\ref{apx_sec:ood_generalization}, we find the consistent OOD generalization performance of {\MPN}  when {\PEN}-based   adversarial perturbations are used in {\MPN}.
In Fig.\,\ref{fig: diff_arch_revised} of Appendix\,\ref{apx_sec:mpn_for_at}, we peer into the generalization of {\MPN} across various {\VM} architectures (\textit{i.e.}, {\AT} in Table\,\ref{tab: attacks}), while maintaining configurations for other attributes  ({\KS}, {\AF}, and {\WS}). 
The observed performance is consistent with Fig.\,\ref{fig: id_vs_ood_revised}.





\begin{figure}[t]
\centerline{
\begin{tabular}{cc}
\hspace*{0mm}\includegraphics[width=.25\textwidth,height=!]{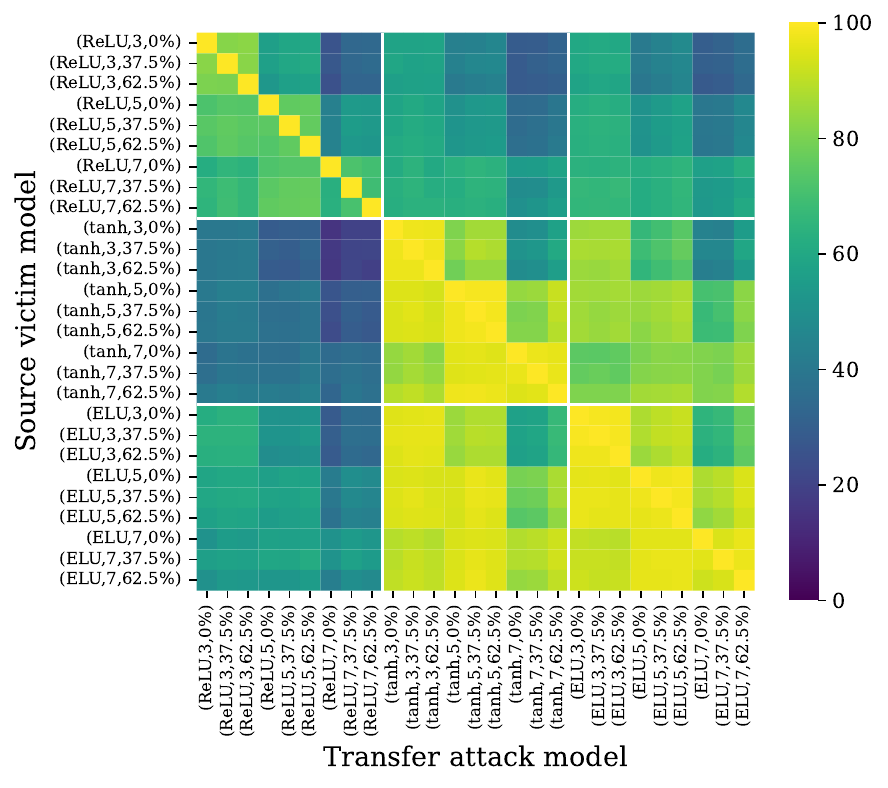}
&\hspace*{-5mm}\includegraphics[width=.25\textwidth,height=!]{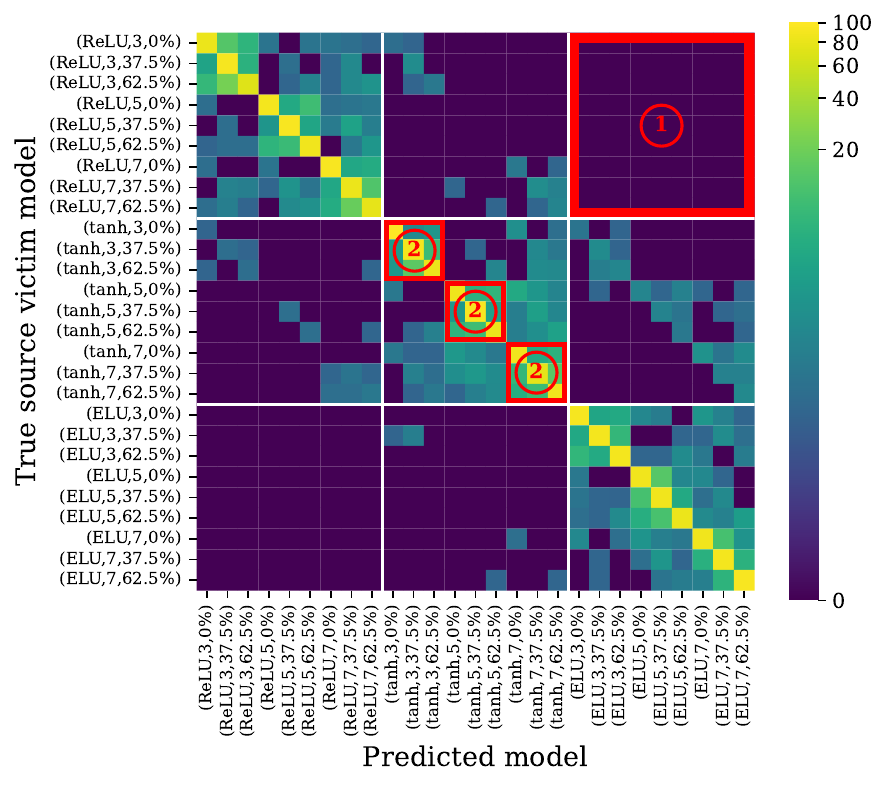}\vspace*{-2mm}\\
 \footnotesize{(a) Attack successful rate (\%)}& \footnotesize{(b) Confusion matrix (\%)}\\
\end{tabular}}
\vspace*{-3mm}
\caption{\footnotesize{Model parsing of transfer attacks: Transfer attack success rate matrix (a) and model parsing confusion matrix (b). Given the architecture type ResNet9, the dataset {\CIFAR}, and the attack type {\PGD} $\ell_\infty$ (with  strength $\epsilon = 8/255$), each   model attribute combination ({\AF}, {\KS}, {\WS}) defines a model instance to be attacked, transferred, or parsed.
}}
 \vspace*{-5mm}
\label{fig: parsing_vs_transfer}
\end{figure}

\noindent 
\textbf{{\MPN} to uncover   real {\VM} attributes of transfer attacks.}
As a use case of model parsing, we next investigate if   {\MPN} can correctly infer the source {\VM} attributes from transfer attacks when  applied to attacking a different model as shown in  Fig.\,\ref{fig: transfer_atk_illustration}.
Given the {\VM} architecture ResNet9, we vary the values of  model attributes  {\KS}, {\AF}, and {\WS}    to produce $27$ ResNet9-type {\VM}s. 
\textbf{Fig.\,\ref{fig: parsing_vs_transfer}} shows the transfer attack success rate (ASR) matrix (Fig.\,\ref{fig: parsing_vs_transfer}a)  and the model parsing confusion matrix (Fig.\,\ref{fig: parsing_vs_transfer}b). Here the transfer attack type is given by   {\PGD} $\ell_\infty $ attack with strength $\epsilon = 8/255$ on {\CIFAR}.


In \textbf{Fig.\,\ref{fig: parsing_vs_transfer}a}, the off-diagonal entries  denote ASRs of transfer attacks   from   row-wise {\VM}s  to attacking   column-wise target models.  Adversarial attacks generated from  ReLU-based {\VM}s are  typically more  difficult to transfer to smooth activation function (ELU or tanh)-based target models. By contrast, given the values of {\AF} and {\KS}, attacks are easier to transfer across models with different weight sparsity. 

 \textbf{Fig.\,\ref{fig: parsing_vs_transfer}b} presents the confusion matrix  of {\MPN} trained on attack data generated from all $27$ ResNet9-alike {\VM}s.
Each row of the confusion matrix represents the true {\VM} used to generate the   attack dataset, and each column  corresponds to a predicted model attribute configuration. Thus,  the diagonal entries and the off-diagonal entries   in Fig.\,\ref{fig: parsing_vs_transfer}b   represent the correct model parsing  accuracy and the misclassification rate on the incorrectly predicted model attribute configuration.  As we can see, attacks  generated from ReLU-based {\VM}s result in a low misclassification rate of {\MPN} on ELU or tanh-based predictions (see the marked region \ding{172}). Meanwhile, a high misclassification occurs for {\MPN} when evaluated on attack data corresponding to different values of {\WS} (see the marked region \ding{173}). The above results, together with our insights into ASRs of transfer attacks in Fig.\,\ref{fig: parsing_vs_transfer}a, suggest a connection between transfer attack and model parsing: \textit{If   attacks are difficult (or easy) to transfer from the source model to the target model, then inferring the source model attributes from these attacks turns to be easy (or difficult).} 
To elucidate the above phenomenon,
our investigation extends to assessing attack transferability via input gradient correlation,  which indicates that a high alignment of gradients between models enhances transferability \cite{demontis2019adversarial}. As seen in Fig.\,\ref{fig: grad_corr} of Appendix\,\ref{apx_sec:transfer_parsing}, we show that attacks which easily transfer across models are difficult to parse accurately due to their gradient features lacking distinctiveness.


\begin{wraptable}{r}{50mm}
\centering
\caption{\footnotesize{Peformance (RA and SA) of model parsing-enabled adversarial defense against transfer attacks. Each row represents either no defense or a defense strategy involving the alteration of attacked model attributes (KS, AF, WS) to differ from the source model attributes used to generate transfer attacks. \greencheck (\redcross) denotes w/ (w/o) modification. The transfer attack configurations follow Fig.\,\ref{fig: parsing_vs_transfer}.
}}
\vspace*{-2mm}
\label{tab: defense}
\resizebox{48mm}{!}{%
\begin{tabular}{c|ccc|c|c}
\toprule[1pt]
\midrule
Setting&\ $\KS$\ & \ $\AF$\  & \ $\WS$\ \  &\  RA (\%) & \ SA (\%)\\ 
\midrule

No Defense&
 \redcross
&  \redcross
&  \redcross
& 0
& 90.7

\\
\midrule
 &
 \greencheck
&  \redcross
&  \redcross
& 31.2
& 90.4

\\
Change 1 Attr. &
 \redcross
&  \greencheck
&  \redcross
& 50.1
& 91.7
\\

 &
 \redcross
& \redcross
&  \greencheck
& 10.9
& 90.6

\\
\midrule

 &
 \greencheck
&  \greencheck
& \redcross
& 63.3
& 91.1

\\

Change 2 Attr.&
\greencheck
& \redcross
& \greencheck
& 34.6
& 90.1

\\

&
 \redcross
&  \greencheck
& \greencheck
& 51.5
& 91.8

\\
\midrule

Change 3 Attr. &
\greencheck
& \greencheck
& \greencheck
& \textbf{66.3}
& 91.2
\\


\midrule
\bottomrule[1pt]
\end{tabular}%
}
\vspace*{-4mm}
\end{wraptable}
\noindent \textbf{Defense inspired by model parsing.} 
Inspired by the {\MPN}'s ability to infer source model attributes of transfer attacks (Fig.\,\ref{fig: parsing_vs_transfer}), we propose an adversarial defense scheme. This scheme alters the target model's attributes to differ from those of the source model, improving its robustness against transfer attacks. Our rationale is that a model can achieve improved robustness against transfer attacks if it has distinct attributes from the source model used to generate these attacks. 
\textbf{Fig.\,\ref{fig: parsing_vs_transfer}} shows the robust accuracy (RA) and standard accuracy (SA) of the above defense method. As we can see, without any defense, transfer attacks remain effective. However, robustness (measured by RA) increases when one attribute is modified, especially when altering the {\AF} type. This is expected, as modifying activation functions has been shown to improve model robustness \cite{xie2020smooth}. Furthermore, when all attributes are modifiable, the defense achieves the highest RA without compromising SA.



\noindent \textbf{Other experimental studies.}
We examine the influence of adversarial robustness of {\VM}s in  the generalization performance of {\MPN}; see Fig.\,\ref{fig: robust_parsing_revised} in {Appendix\,\ref{apx_sec:parsing_vs_robustness}.}
We find that adversarial attacks against robust {\VM}s is harder to parse {\VM} attributes than attacks against standard {\VM}s. 
The possible rationale is that attacks carry rich {\VM} information in input gradients. Yet, robust models typically exhibit weaker input gradient norms compared to standard models \cite{finlay2021scaleable}. Thus, robust models are harder to infer due to the low power of input gradients. Fig.\,\ref{fig: grad_norm} provides an empirical justification for this rationale.
In Appendix\,\ref{apx_sec:ablation}, we examine the influence of other factors like PGD steps, step sizes, and transfer attack strengths on the generalization performance of {\MPN}.




%% file: secs/conclusion.tex
 \vspace*{-3mm}
\section{Conclusion}
 \vspace*{-2mm}

We study model parsing from adversarial attacks to deduce attributes of victim models, with the development of model parsing network ({\MPN}). Our exploration spanned both in-distribution and out-of-distribution scenarios, evaluating {\MPN} against diverse attack methods and model configurations. Key determinants such as input format, backbone network, and attack characteristics were analyzed for their impact on model parsing. We elucidated the conditions under which victim model information can be extracted from adversarial attacks. Our study  empowers defenders with an enhanced understanding of attack provenance.



%% file: secs/acknowledgement.tex
\section{Acknowledgement}
The work was supported by the DARPA RED program.

%% file: secs/appendix.tex
\appendix
\onecolumn
\setcounter{table}{0}  
\setcounter{figure}{0}

\renewcommand{\thetable}{A\arabic{table}}
\renewcommand{\thefigure}{A\arabic{figure}}

\begin{center}
    \Large{\textbf{{Appendix}}}
\end{center}

\section{Victim model training, evaluation, and attack setups} 
\label{apx_sec:victim_model_train}

When training all {\CIFAR}, {\CIFARM}, and {\TImageNet} victim models (each of which is given by an   attribute combination), we use the SGD optimizer with the cosine annealing learning rate schedule and an initial learning rate of 0.1. The weight decay is $5e-4$, and the batch size is 256. The number of training epochs is 75 for {\CIFAR} and {\CIFARM}, and  100 for {\TImageNet}. 
When the weight sparsity ({\WS}) is promoted,  we follow the one-shot magnitude pruning method \cite{frankle2018lottery,ma2021sanity} to 
 obtain a sparse model.
To obtain models with different activation functions ({\AF}) and kernel sizes ({\KS}), we modify the convolutional block design in the ResNet and VGG model family accordingly from 3, ReLU to   others, \textit{i.e.,} 5/7, tanh/ELU.
Table\,\ref{tab: victim_model} shows the testing accuracy (\%) of victim models on different datasets, given any studied ({\AF}, {\KS}, {\WS}) tuple included in Table\,\ref{tab: models}. It is worth noting that we accelerate victim model training by using FFCV \cite{leclerc2023ffcv} when loading the dataset. 

\begin{table*}[ht]
\centering
\caption{\footnotesize{
Victim model performance (testing accuracy, \%) given different choices of datasets and model architectures. 
}}
\label{tab: victim_model}
\resizebox{\textwidth}{!}{
\begin{tabular}{c|c|c|c|c|c|c|c|c|c|c|c|c|c|c|c|c|c|c|c|c|c|c|c|c|c|c|c|c}
\toprule[1pt]\midrule
\multirow{4}{*}{\begin{tabular}[c]{@{}c@{}}\textbf{Dataset}\end{tabular}} & 
\multirow{4}{*}{\begin{tabular}[c]{@{}c@{}} \AT\end{tabular}}
& \multicolumn{27}{c}{\textbf{Attribute combination}} \\
 & & \multicolumn{9}{c|}{ReLU} & \multicolumn{9}{c|}{tanh} & \multicolumn{9}{c}{ELU}
\\
& & \multicolumn{3}{c|}{3} & \multicolumn{3}{c|}{5} & \multicolumn{3}{c|}{7} & \multicolumn{3}{c|}{3} & \multicolumn{3}{c|}{5} & \multicolumn{3}{c|}{7} & \multicolumn{3}{c|}{3} & \multicolumn{3}{c|}{5} & \multicolumn{3}{c}{7} \\
& & 0\% & 37.5\% & 62.5\% & 0\% & 37.5\% & 62.5\% & 0\% & 37.5\% & 62.5\% & 0\% & 37.5\% & 62.5\% & 0\% & 37.5\% & 62.5\% & 0\% & 37.5\% & 62.5\% & 0\% & 37.5\% & 62.5\% & 0\% & 37.5\% & 62.5\% & 0\% & 37.5\% & 62.5\% \\
\midrule
\multirow{5}{*}{\begin{tabular}[c]{@{}c@{}}{\CIFAR}\end{tabular}} 
& ResNet9 
        &94.4   &93.9   &94.2   &93.3   &93.5   &93.5   &92.4   &92.8   &92.8   &89.0   &88.8   &89.9   &88.4   &88.6   &88.2   &87.0   &87.2   &88.0   &91.0   &91.2   &90.7   &90.3   &90.2   &90.5   &89.3   &90.0   &89.6\\
        & ResNet18
        &94.7   &94.9   &95.0   &94.2   &94.5   &94.5   &93.9   &93.5   &93.6   &87.1   &87.5   &88.2   &84.2   &84.9   &85.5   &81.3   &81.2   &85.1   &90.6   &90.8   &90.6   &90.1   &91.1   &90.5   &85.7   &83.4   &84.3\\
        & ResNet20
        &92.1   &92.5   &92.3   &92.0   &92.2   &92.0   &90.9   &91.8   &91.5   &89.7   &89.7   &89.7   &89.5   &89.4   &89.6   &88.3   &88.2   &88.9   &90.7   &91.2   &90.9   &90.3   &90.5   &90.6   &89.2   &89.7   &89.7\\
        & VGG11
        &91.0   &91.1   &90.4   &89.8   &89.9   &89.4   &88.2   &88.4   &88.0   &88.7   &89.1   &88.9   &87.2   &87.6   &87.6   &87.0   &86.8   &87.0   &89.4   &89.5   &89.5   &88.0   &88.2   &88.5   &87.1   &87.1   &87.2\\
        & VGG13
        &93.1   &93.3   &93.0   &92.0   &92.2   &92.6   &91.2   &91.1   &91.0   &90.1   &90.1   &90.1   &89.3   &89.1   &89.3   &88.2   &88.8   &88.8   &90.8   &90.9   &90.8   &89.2   &89.5   &89.4   &88.4   &88.7   &88.9\\
        \midrule
\multirow{5}{*}{{\CIFARM}} 
& ResNet9
        &73.3   &73.6   &73.5   &71.8   &71.9   &71.2   &69.1   &69.8   &69.2   &58.6   &60.1   &60.3   &60.1   &61.2   &62.0   &58.2   &59.8   &60.3   &70.8   &70.7   &70.8   &69.5   &69.6   &69.8   &67.3   &68.3   &68.7\\
        & ResNet18       &74.4   &75.0   &75.6   &73.6   &73.0   &74.6   &71.2   &71.0   &70.9   &62.0   &62.0   &62.9   &57.3   &59.3   &60.1   &51.3   &53.4   &57.1   &70.1   &70.8   &71.1&66.8    &69.7   &69.7   &63.1   &61.8   &65.7\\
        &  ResNet20       &68.3   &68.4   &67.5   &67.8   &67.5   &67.7   &66.8   &66.7   &67.6   &59.9   &61.3   &59.6   &61.9   &62.0   &62.1   &59.9   &61.2   &61.2   &66.4   &67.6   &67.7&67.0    &67.3   &67.2   &66.2   &66.9   &66.8\\
        
       &  VGG11         &68.3   &68.4   &67.7   &65.2   &65.7   &65.8   &62.4   &62.0   &62.6   &65.2   &65.5   &65.5   &63.6   &63.6   &63.9   &62.1   &61.8   &62.5   &66.2   &66.5   &65.9&64.6    &64.0   &64.6   &61.5   &62.3   &61.9\\
        
       &  VGG13         &71.0   &70.6   &71.1   &69.9   &70.5   &70.3   &66.5   &66.5   &67.2   &66.7   &67.5   &67.5   &65.2   &65.5   &67.1   &63.9   &63.4   &65.0   &68.9   &69.3   &69.5&66.3    &66.7   &67.1   &64.2   &64.5   &64.7\\

        \midrule
\multirow{1}{*}{{\TImageNet}} 
& ResNet18
        &63.7   &64.1   &63.5   &61.5   &62.7   &62.6   &59.6   &61.0   &61.7   &47.0   &48.1   &50.0   &46.6   &47.9   &48.3   &41.0   &43.5   &44.6   &57.2   &57.9   &58.1   &52.7   &53.8   &53.6   &52.3   &51.5   &52.3\\
        \midrule\bottomrule[1pt]
\end{tabular}}
\vspace{-3mm}
\end{table*}

For different attack types, we list all the attack configurations below:

\noindent \ding{70} {\FGSM}. We set the attack strength $\epsilon$ equal to $4/255$, $8/255$, $12/255$, and $16/255$, respectively.

\noindent \ding{70} {\PGD} $\ell_\infty$. We set the attack step number equal to 10, and the attack strength-learning rate combinations as  ($\epsilon = 4/255$, $\alpha = 0.5/255$),  ($\epsilon = 8/255$, $\alpha=1/255$),  ($\epsilon = 12/255$, $\alpha = 2/255$), and  ($\epsilon = 16/255$, $\alpha=2/255$).


\noindent \ding{70} {\PGD} $\ell_2$. We set the step number equal to 10, and the attack strength-learning rate combinations as ($\epsilon = 0.25$, $\alpha = 0.05$), ($\epsilon = 0.5$, $\alpha = 0.1$), ($\epsilon = 0.75$, $\alpha = 0.15$), and ($\epsilon = 1.0$, $\alpha = 0.2$).

\noindent \ding{70} {\CW}. We use $\ell_2$ version CW attack with the attack conference parameter  $\kappa$ equal to $0$.
We also set the learning rate equal to $0.01$ and  the maximum iteration number equal to $50$ to search for successful attacks.

\noindent \ding{70} {\AutoAttack} $\ell_\infty$. We use the standard version of {\AutoAttack} with the $\ell_\infty$ norm and  $\epsilon$ equal to $4/255$, $8/255$, $12/255$, and $16/255$, respectively.

\noindent \ding{70} {\AutoAttack} $\ell_2$. We use the standard version of {\AutoAttack} with the $\ell_2$ norm  and  $\epsilon$ equal to $0.25$, $0.5$, $0.75$, and $1.0$, respectively.

\noindent \ding{70} {\SquareAttack} $\ell_\infty$. We set the maximum query number equal to $5000$ with $\ell_\infty$ norm $\epsilon$ equal to $4/255$, $8/255$, $12/255$, and $16/255$, respectively.

\noindent \ding{70} {\SquareAttack} $\ell_2$.  We set the maximum query number equal to $5000$ with $\ell_\infty$ norm $\epsilon$ equal to $0.25$, $0.5$, $0.75$, and $1.0$, respectively.

\noindent \ding{70} {\NES}. We set the query number for each gradient estimate equal to 10, together with $\mu = 0.01$ (\textit{i.e.,} the value of the smoothing parameter to obtain the finite difference of function evaluations). We also set the learning rate by $0.0005$, and the maximum iteration number by 500 for each adversarial example generation.

\noindent \ding{70} {\ZOSignSGD}.  We set the query number for each gradient estimate equal to 10 with $\mu = 0.01$.  We also set the learning rate equal to $0.0005$,  and the maximum iteration number equal to 500 for each adversarial example generation.
The only difference between {\ZOSignSGD} and {\NES} is the gradient estimation method in ZOO. {\ZOSignSGD} uses the sign of forward difference-based estimator while  {\NES} uses the central difference-based estimator.

\clearpage
\newpage
\section{OOD generalization performance of {\MPN} across attack types when {\PEN} is used} \label{apx_sec:ood_generalization}
\begin{figure}[htb]
\centerline{
\includegraphics[width=0.7\textwidth]{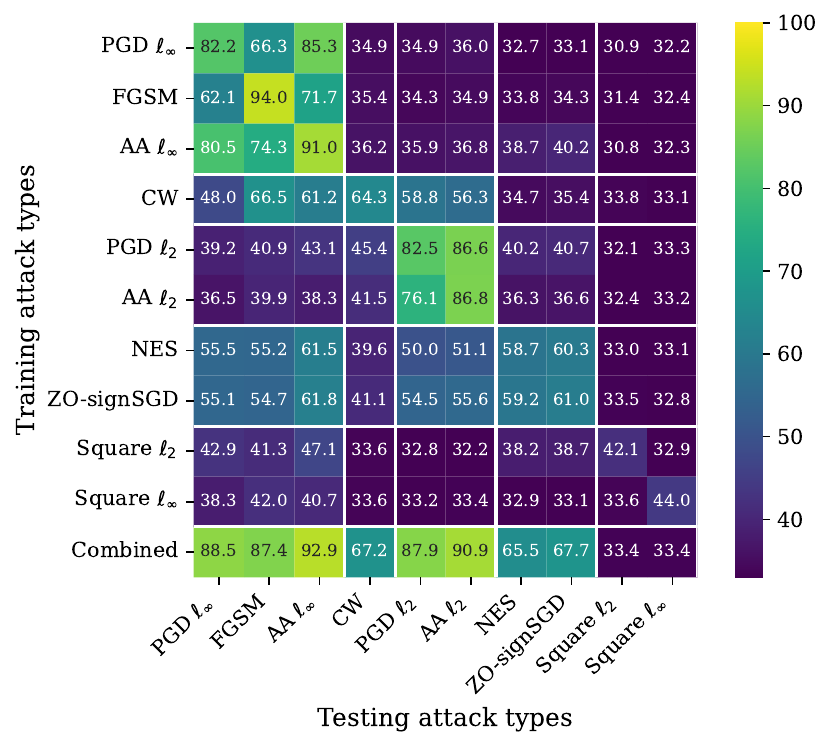}
}
\caption{\footnotesize{Generalization performance matrix of {\MPN} when trained on  a row-specific attack type but evaluated on a column-specific attack type given $\bdelta_{\PEN}$ as input. The attack data are given by {adversarial perturbations} with strength  $\epsilon = 8/255$ for $\ell_\infty$ attacks, $\epsilon = 0.5$ for $\ell_2$ attacks, and $c=1$ for {\CW} attack. The victim model architecture and the dataset are set as ResNet9 and {\CIFAR}. The `combined' row represents {\MPN} training on the collection of four attack types: {\PGD} $\ell_\infty$, {\PGD} $\ell_2$, {\CW}, and {\ZOSignSGD}. 
}}
 \vspace*{-1mm}
\label{fig: denoise_ood}
\end{figure}

Similar to Fig.\,\ref{fig: parsing_vs_transfer}, Fig.\,\ref{fig: denoise_ood} shows the generalization performance of {\MPN} when trained on a row-specific attack type but evaluated on a column-specific attack type when $\bdelta_\PEN$ is given as input. When {\MPN} is trained on the collection of four attack types {\PGD} $\ell_\infty$, {\PGD} $\ell_2$, {\CW}, and {\ZOSignSGD} (\textit{i.e.}, the `Combined' row), such a data augmentation can boost the OOD generalization except for the random search-based {\SquareAttackShort} attack.

\clearpage
\newpage
\section{{\MPN} across different architecture types ({\AT})} 
\label{apx_sec:mpn_for_at}

We also peer into the generalization of {\MPN} across different {\VM} architectures (\textit{i.e.}, {\AT} in Table\,\ref{tab: attacks}), while maintaining constant configurations for other attributes  ({\KS}, {\AF}, and {\WS}). 
 
\begin{figure}[htb]
\vspace*{-1mm}
\centerline{
\begin{tabular}{ccc}
\hspace*{2mm}\includegraphics[width=.3\textwidth,height=!]{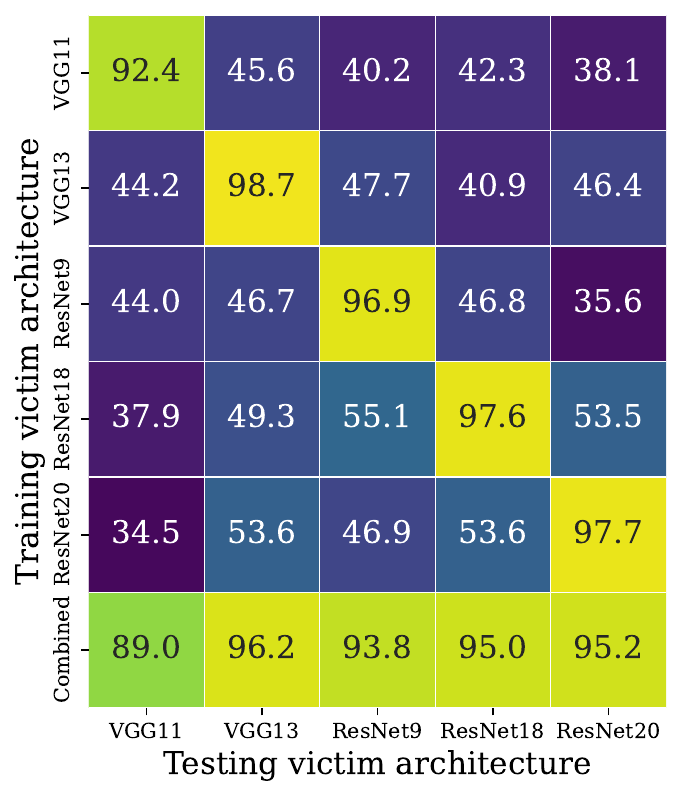}
&\hspace*{0mm}\includegraphics[width=.3\textwidth,height=!]{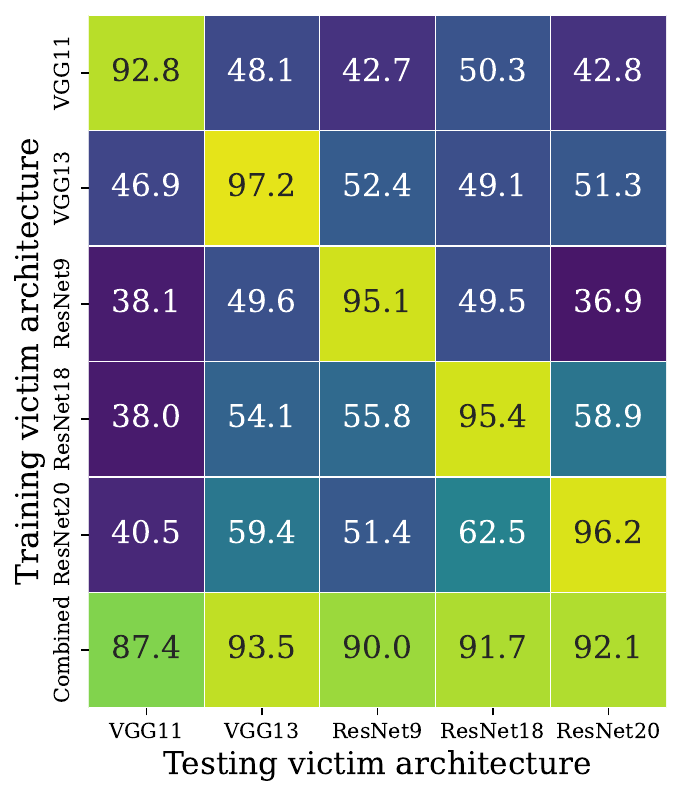}
&\hspace*{0mm}\includegraphics[width=.356\textwidth,height=!]{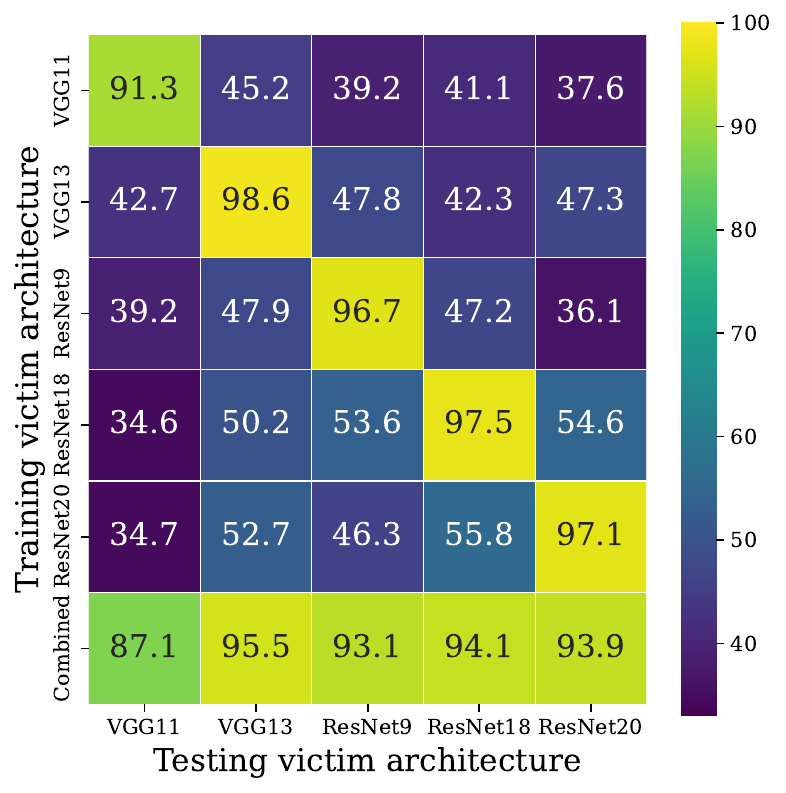}\\
 \hspace*{2mm}\footnotesize{(a) {\FGSM}} &\hspace*{-4mm} \footnotesize{(b) {\PGD} $\ell_\infty$} &\hspace*{-4mm} \footnotesize{(c)  {\CW}}\\
\end{tabular}
}
\vspace{-3mm}
\caption{\footnotesize{
Generalization   matrix (\%)  of {\MPN} when trained on attack data generated from a
row-specific architecture  but evaluated on attack data generated from  a column-specific architecture. Both the train-time and   test-time 
 architectures share the same {\VM} attributes in   {\KS}, {\AF}, and {\WS}. The   attack type is specified by   {\FGSM}, {\PGD} $\ell_\infty$, or {\CW} on {\CIFAR}, with the attack strength $\epsilon = 8/255$ for $\ell_\infty$ attacks and $c = 1$ for {\CW}.
}}
\vspace*{-3mm}
\label{fig: diff_arch_revised}
\end{figure}

\textbf{Fig.\,\ref{fig: diff_arch_revised}} demonstrates the generalization matrix of {\MPN} when trained and evaluated using adversarial perturbations   generated from different {\VM} architectures  (\textit{i.e.}, different values of {\AT} in Table\,\ref{tab: attacks}) by fixing the configurations    of other attributes ({\KS}, {\AF}, and {\WS}).
We observe that given an attack type, the in-distribution {\MPN} generalization  remains well across {\VM} architectures. Yet, the OOD generalization of {\MPN} (corresponding to the off-diagonal entries of the generalization matrix) rapidly degrades if the test-time  {\VM} architecture  is different from the train-time one. This inspires us to train {\MPN} on more {\AT} variants in order to retain the model parsing performance, as shown in the last row of each subfigure of Fig.\,\ref{fig: diff_arch_revised}.

{\MPN} is then trained on ({\AT}, {\AF}, {\KS}, {\WS}) tuple by merging {\AT} into the attribute classification task. We conduct experiments considering different architectures mentioned in Table\,\ref{tab: models} on {\CIFAR} and {\CIFARM}, with   $\bdelta$ and $\bdelta_\PEN$ as {\MPN}'s inputs, respectively.
We summarize the in-distribution generalization results
  in Table\,\ref{tab: MPN_acc}, Table\,\ref{tab: MPN_acc_PEN}, Table\,\ref{tab: MPN_acc_100}, and Table\,\ref{tab: MPN_acc_PEN_100}. Weighted accuracy refers to the testing accuracy defined in Sec.\,\ref{sec: exp}, \textit{i.e.,} $\sum_i (N_i \mathrm{TA} (i) ) / \sum_i N_i$, where $N_i$ is the number of classes of the model attribute $i$, and $ \mathrm{TA} (i)$ is the testing accuracy of  the classifier associated with the attribute $i$ (Fig.\,\ref{fig: overview_classifier}).
  In the above tables, we  also show  the testing accuracy for each attribute, \textit{i.e.,} $ \mathrm{TA} (i)$.
  Combined accuracy refers to the testing accuracy over all victim model attribute-combined classes, \textit{i.e.}, 135 classes for 5 {\AT} classes, 3 {\AF} classes, 3 {\KS} classes, and 3 {\WS} classes. 
 The insights into model parsing are summarized below: (1) {\MPN} trained on $\bdelta$ and $\bdelta_{\PEN}$ can effectively classify all the attributes  {\AT}, {\AF}, {\KS}, {\WS} in terms of per-attribute classification accuracy, weighted testing accuracy, and combined accuracy. (2)  Compared to {\AT}, {\AF}, and {\KS}, {\WS} is harder to parse.

\begin{table*}[ht]
\centering
\caption{\footnotesize{
{\MPN} performance (\%) on different attack types given different evaluation metrics with adversarial perturbation $\bdelta$ as input on {\CIFAR}.
}}
\label{tab: MPN_acc}
\vspace{1mm}
\resizebox{\textwidth}{!}{
\begin{tabular}{c|c|c|c|c|c|c|c|c|c|c|c|c|c|c|c}
\toprule[1pt]\midrule
\multirow{3}{*}{\begin{tabular}[c]{@{}c@{}}\textbf{Metrics}\end{tabular}} & 
 \multicolumn{15}{c}{\textbf{Attack types}} \\
& \multicolumn{4}{c|}{{\FGSM}} &  \multicolumn{4}{c|}{{\PGD} $\ell_\infty$} & \multicolumn{4}{c|}{{\PGD} $\ell_2$} & \multicolumn{3}{c}{\CW}
\\
& $\epsilon=4/255$ & $\epsilon=8/255$ & $\epsilon=12/255$ & $\epsilon=16/255$ & $\epsilon=4/255$ & $\epsilon=8/255$ & $\epsilon=12/255$ & $\epsilon=16/255$ & $\epsilon=0.25$ & $\epsilon=0.5$ & $\epsilon=0.75$ & $\epsilon=1.0$ & $c=0.1$ & $c=1$ & $c=10$  \\
\midrule
{\AT} accuracy                      &97.77  &97.85  &97.91  &97.91  &97.23  &96.13  &96.16  &94.22  &99.77  &99.64  &99.37  &99.12  &96.73  &97.30  &97.28
\\
{\AF} accuracy              &95.67  &95.73  &95.79  &95.71  &95.86  &95.26  &95.77  &94.05  &99.51  &99.36  &99.04  &98.68  &95.12  &94.84  &94.68
\\
{\KS} accuracy           &98.66  &98.66  &98.65  &98.71  &98.22  &97.55  &97.43  &95.52  &99.83  &99.79  &99.64  &99.48  &96.94  &98.13  &98.09
\\
{\WS} accuracy                &87.16  &87.16  &87.29  &87.52  &84.36  &79.99  &80.01  &71.68  &98.51  &97.83  &96.86  &95.57  &88.42  &85.28  &85.03
\\ 
        \midrule
Weighted accuracy                &95.24  &95.28  &95.34  &95.38  &94.39  &92.79  &92.89  &89.63  &99.46  &99.23  &98.82  &98.34  &94.65  &94.38  &94.27
\\ 
        \midrule
Combined accuracy              &81.85  &82.00  &82.19  &82.33  &78.65  &73.11  &73.33  &62.67  &97.79  &96.89  &95.38  &93.55  &83.00  &79.29  &78.88
\\ 
        \midrule\bottomrule[1pt]
\end{tabular}}
\end{table*}

\begin{table*}[ht]
\centering
\caption{\footnotesize{
{\MPN} performance (\%) on different attack types given different  evaluation metrics with estimated perturbation $\bdelta_\PEN$ as input on {\CIFAR}.
}}
\label{tab: MPN_acc_PEN}
\vspace{1mm}
\resizebox{\textwidth}{!}{
\begin{tabular}{c|c|c|c|c|c|c|c|c|c|c|c|c|c|c|c}
\toprule[1pt]\midrule
\multirow{3}{*}{\begin{tabular}[c]{@{}c@{}}\textbf{Metrics}\end{tabular}} & 
 \multicolumn{15}{c}{\textbf{Attack types}} \\
& \multicolumn{4}{c|}{{\FGSM}} &  \multicolumn{4}{c|}{{\PGD} $\ell_\infty$} & \multicolumn{4}{c|}{{\PGD} $\ell_2$} & \multicolumn{3}{c}{\CW}
\\
& $\epsilon=4/255$ & $\epsilon=8/255$ & $\epsilon=12/255$ & $\epsilon=16/255$ & $\epsilon=4/255$ & $\epsilon=8/255$ & $\epsilon=12/255$ & $\epsilon=16/255$ & $\epsilon=0.25$ & $\epsilon=0.5$ & $\epsilon=0.75$ & $\epsilon=1.0$ & $c=0.1$ & $c=1$ & $c=10$  \\
\midrule
{\AT} accuracy                   &88.98  &95.68  &97.20  &97.64  &75.81  &84.58  &90.27  &88.50  &61.09  &81.41  &87.80  &90.48  &56.10  &64.11  &64.30
\\
{\AF} accuracy                   &83.48  &92.21  &94.56  &95.22  &74.95  &85.04  &90.72  &89.81  &57.62  &76.90  &83.95  &87.36  &54.61  &58.77  &58.98
\\
{\KS} accuracy                &91.57  &96.63  &97.96  &98.41  &81.10  &88.18  &92.67  &90.99  &67.85  &84.50  &89.93  &92.18  &62.46  &69.81  &70.15
\\
{\WS} accuracy                 &69.99  &81.42  &84.92  &86.59  &56.07  &63.92  &70.19  &64.80  &50.09  &67.26  &74.02  &77.70  &46.40  &47.53  &47.77
\\ 
        \midrule
Weighted accuracy               &84.29  &92.08  &94.17  &94.92  &72.53  &81.02  &86.58  &84.23  &59.44  &78.07  &84.48  &87.44  &55.07  &60.63  &60.87
\\ 
        \midrule
Combined accuracy               &54.83  &72.66  &78.63  &80.83  &32.59  &46.05  &57.10  &50.60  &18.38  &45.39  &57.10  &63.00  &14.62  &19.44  &19.70
\\ 
        \midrule\bottomrule[1pt]
\end{tabular}}
\end{table*}

\begin{table*}[ht]
\centering
\caption{\footnotesize{
{\MPN} performance (\%) on different attack types given different evaluation metrics with adversarial perturbation $\bdelta$ as input on {\CIFARM}.
}}
\label{tab: MPN_acc_100}
\vspace{1mm}
\resizebox{\textwidth}{!}{
\begin{tabular}{c|c|c|c|c|c|c|c|c|c|c|c|c|c|c|c}
\toprule[1pt]\midrule
\multirow{3}{*}{\begin{tabular}[c]{@{}c@{}}\textbf{Metrics}\end{tabular}} & 
 \multicolumn{15}{c}{\textbf{Attack types}} \\
& \multicolumn{4}{c|}{{\FGSM}} &  \multicolumn{4}{c|}{{\PGD} $\ell_\infty$} & \multicolumn{4}{c|}{{\PGD} $\ell_2$} & \multicolumn{3}{c}{\CW}
\\
& $\epsilon=4/255$ & $\epsilon=8/255$ & $\epsilon=12/255$ & $\epsilon=16/255$ & $\epsilon=4/255$ & $\epsilon=8/255$ & $\epsilon=12/255$ & $\epsilon=16/255$ & $\epsilon=0.25$ & $\epsilon=0.5$ & $\epsilon=0.75$ & $\epsilon=1.0$ & $c=0.1$ & $c=1$ & $c=10$  \\
\midrule
{\AT} accuracy               &97.70  &97.76  &97.76  &97.75  &97.03  &95.40  &95.23  &92.52  &99.59  &99.29  &98.91  &98.50  &93.84  &96.23  &96.30
\\
{\AF} accuracy              &95.17  &95.14  &94.96  &95.11  &94.79  &93.73  &93.87  &91.87  &99.14  &98.63  &97.97  &97.31  &90.83  &92.32  &92.47
\\
{\KS} accuracy               &97.66  &97.65  &97.69  &97.62  &96.75  &95.16  &94.44  &91.25  &99.62  &99.43  &99.16  &98.70  &93.11  &95.77  &95.81
\\
{\WS} accuracy         &81.13  &80.77  &80.90  &80.94  &76.57  &69.85  &68.16  &59.42  &96.58  &95.04  &92.70  &90.43  &76.61  &74.64  &74.77
\\ 
        \midrule
Weighted accuracy         &93.60  &93.54  &93.53  &93.55  &92.11  &89.52  &88.97  &85.02  &98.85  &98.27  &97.43  &96.56  &89.34  &90.67  &90.76
\\ 
        \midrule
Combined accuracy         &75.08  &74.76  &74.82  &74.95  &69.72  &61.27  &59.31  &48.37  &95.27  &93.06  &89.89  &86.73  &67.19  &66.24  &66.56
\\ 
        \midrule\bottomrule[1pt]
\end{tabular}}
\end{table*}

\begin{table*}[ht]
\centering
\caption{\footnotesize{
{\MPN} performance (\%) on different attack types given different evaluation metrics with estimated perturbation $\bdelta_\PEN$ as input on {\CIFARM}.
}}
\label{tab: MPN_acc_PEN_100}
\vspace{1mm}
\resizebox{\textwidth}{!}{
\begin{tabular}{c|c|c|c|c|c|c|c|c|c|c|c|c|c|c|c}
\toprule[1pt]\midrule
\multirow{3}{*}{\begin{tabular}[c]{@{}c@{}}\textbf{Metrics}\end{tabular}} & 
 \multicolumn{15}{c}{\textbf{Attack types}} \\
& \multicolumn{4}{c|}{{\FGSM}} &  \multicolumn{4}{c|}{{\PGD} $\ell_\infty$} & \multicolumn{4}{c|}{{\PGD} $\ell_2$} & \multicolumn{3}{c}{\CW}
\\
& $\epsilon=4/255$ & $\epsilon=8/255$ & $\epsilon=12/255$ & $\epsilon=16/255$ & $\epsilon=4/255$ & $\epsilon=8/255$ & $\epsilon=12/255$ & $\epsilon=16/255$ & $\epsilon=0.25$ & $\epsilon=0.5$ & $\epsilon=0.75$ & $\epsilon=1.0$ & $c=0.1$ & $c=1$ & $c=10$  \\
\midrule
{\AT} accuracy        &88.17  &95.25  &96.92  &97.45  &72.40  &82.48  &88.11  &85.47  &62.77  &80.01  &85.88  &88.33  &47.31  &51.80  &52.48
\\
{\AF} accuracy        &81.81  &91.14  &93.53  &94.52  &71.43  &81.93  &87.76  &86.71  &58.16  &74.06  &80.88  &84.18  &49.98  &49.49  &49.96
\\
{\KS} accuracy        &88.62  &94.92  &96.58  &97.12  &76.97  &84.74  &88.78  &86.46  &69.38  &84.09  &88.68  &90.56  &56.07  &59.68  &59.72
\\
{\WS} accuracy         &64.19  &74.98  &78.60  &79.85  &50.64  &56.88  &60.32  &54.94  &46.50  &61.73  &67.79  &70.59  &39.46  &39.85  &40.37
\\ 
        \midrule
Weighted accuracy         &81.76  &89.95  &92.19  &92.98  &68.51  &77.36  &82.22  &79.40  &59.71  &75.69  &81.53  &84.12  &48.08  &50.43  &50.90
\\ 
        \midrule
Combined accuracy         &47.75  &65.27  &71.05  &73.27  &25.56  &37.49  &45.28  &38.97  &16.27  &38.87  &49.04  &54.06  &7.31   &9.20   &9.59
\\ 
        \midrule\bottomrule[1pt]
\end{tabular}}
\end{table*}

\clearpage
\newpage
\section{Correlation between transfer attack and model parsing}
\label{apx_sec:transfer_parsing}

\begin{figure}[htbp]
\centering
\includegraphics[width=0.5\textwidth]{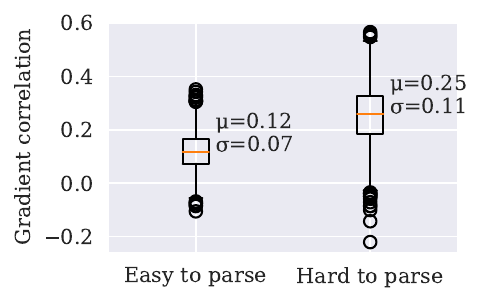}
\caption{Input gradient correlation between true victim model and transfer attack target model in `easy-to-parse' and `hard-to-parse' scenarios, expanded from Fig.\,\ref{fig: parsing_vs_transfer}.}
\label{fig: grad_corr}
\end{figure}

Although model parsing for transfer attacks presents difficulties compared to non-transfer scenarios, our model parsing method does not fail. Fig.\,\ref{fig: parsing_vs_transfer}(b) highlights the dominance of correct model parsing (diagonal entries) over misclassifications (off-diagonal entries). Further, to understand why attack transferability plays a role in model parsing, we extract two representative cases from Fig.\,\ref{fig: parsing_vs_transfer}: (1) `hard-to-parse' (`easy-to-transfer'): transfer attacks from (ReLU,3,0\%) to (ReLU,3,37.5\%); (2) `easy-to-parse' (`hard-to-transfer'): transfer attacks from (ReLU,3,0\%) to (tanh,3,0\%).  Hard-to-parse attacks show stronger input gradient correlation (see \textbf{Fig. \ref{fig: grad_corr}}) between the source victim model and the transfer attack target model, indicating higher transferability. Given the model attribute information is carried within the input gradient, model parsing for transfer attacks is harder.

\clearpage
\newpage

\section{Model parsing vs. model robustness} \label{apx_sec:parsing_vs_robustness}

\begin{figure}[htb]
\centerline{
\includegraphics[width=0.5\textwidth]{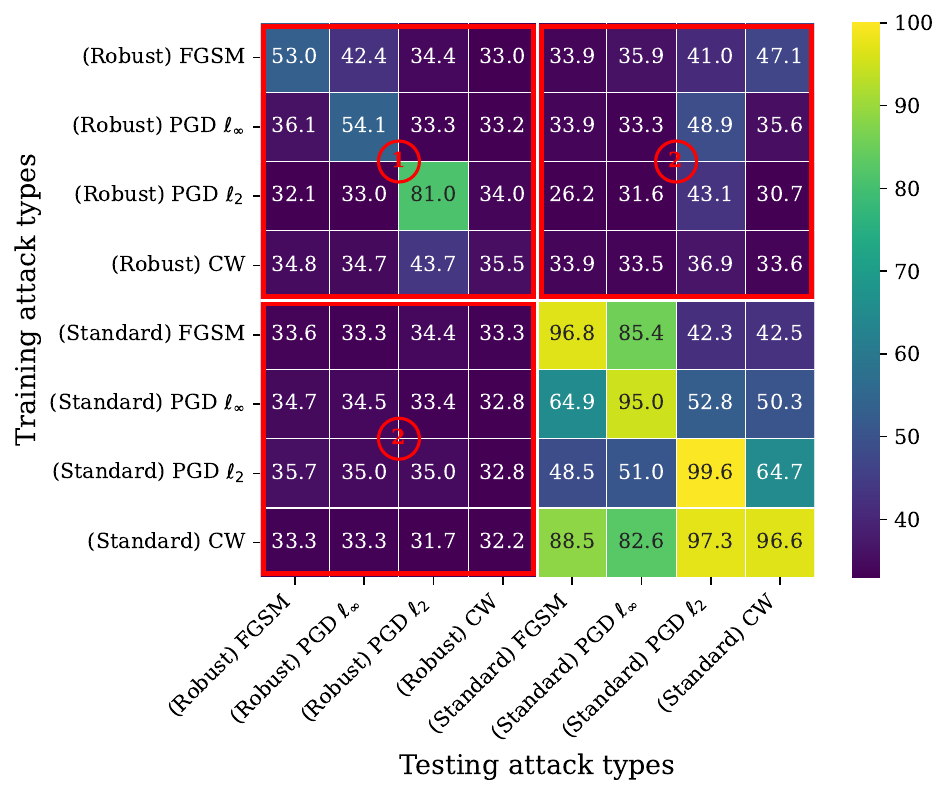}
}
\caption{\footnotesize{
Generalization performance (\%) matrix of {\MPN} across attack types, ranging from {\FGSM}, {\PGD} $\ell_\infty$, {\PGD} $\ell_2$, and {\CW} attacks against standard victim models to their variants against   robust victim models, termed (Standard or Robust) \texttt{Attack}. Other setups are consistent with Fig.\,\ref{fig: id_vs_ood_revised}.
}
}
 \vspace*{-1mm}
\label{fig: robust_parsing_revised}
\end{figure}

We re-use the collected   ResNet9-type victim models in  Fig.\,\ref{fig: parsing_vs_transfer}, and obtain their adversarially robust versions by conducting adversarial training \cite{madry2017towards} on {\CIFAR}. 
Fig.\,\ref{fig: robust_parsing_revised}   presents 
the generalization matrix  of {\MPN} when trained on a row-wise attack type  but evaluated on a column-wise attack type. Yet, different from Fig.\,\ref{fig: id_vs_ood_revised}, the considered attack type is expanded by incorporating `attack against robust  model', besides `attack against standard  model'. It is worth noting that every attack type corresponds to attack data generated from victim models ({\VM}s) instantiated by   the combinations of model attributes {\KS}, {\AF}, and {\WS}.
Thus, the diagonal entries and the off-diagonal entries of the generalization matrix in Fig.\,\ref{fig: robust_parsing_revised} reflect the in-distribution parsing accuracy within an attack type  and the OOD generalization across attack types. Here are two key observations. 
First,  the in-distribution generalization of {\MPN} from attacks against robust {\VM}s is much poorer  (see the marked region \ding{172}), compared to that   from  attacks against standard  {\VM}s. 
Second, the off-diagonal performance shows that {\MPN} trained on attacks against standard {\VM}s is harder to parse model attributes from attacks against robust {\VM}s, and vice versa (see the marked region \ding{173}). Based on the above results, we posit that  model parsing is easier for attacks generated from {\VM}s with higher accuracy and lower robustness.

\begin{figure}[htbp]
\centering
\includegraphics[width=0.4\textwidth]{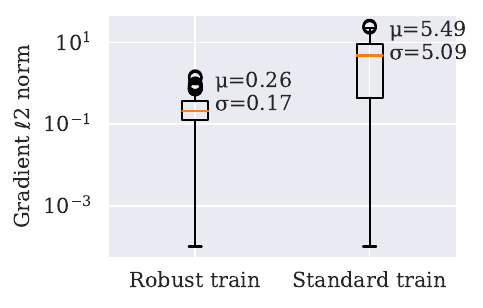}
\caption{Distribution of input gradient $\ell_2$ norms on testing data of robustly and normally trained models on (CIFAR10, ResNet9), expanded from Fig.\,\ref{fig: robust_parsing_revised}.}
\label{fig: grad_norm}
\end{figure}
To explain why the adversarial examples against robust models are hard to parse, we try to understand this from the perspective of the norm of input gradients. 
Fig.\,\ref{fig: grad_norm} shows that robust models are harder to infer from the adversarial examples due to the low magnitude power of input gradients, making it non-differentiable between different model attribute configurations.

\clearpage
\newpage
\section{Other experimental studies}
\label{apx_sec:ablation}
We also study other factors that could possibly affect the model parsing performance like PGD steps, step sizes, and stronger transfer attack methods. 

\begin{table}[th]
\centering
\caption{Model parsing accuracy of PGD $\ell_\infty$ perturbations with $\epsilon=8/255$ on (CIFAR10, ResNet9) under different attack steps $k$ and step sizes $\alpha$.}
\label{tab:PGDstep}
\resizebox{0.5\textwidth}{!}{%
\begin{tabular}{c|c|c|c|c}
\toprule[1pt]
\midrule
\diagbox{\textbf{Training}}{\textbf{Testing}} & 
$k=$ 10 & 20 & 40 & 80 \\
\midrule
$k=$ 10 & 95.07  & 93.59 & 93.53 & 93.53 \\
20 & 93.73  & 93.88 & 93.92 & 93.89 \\
40 & 93.84  & 93.96 & 94.07 & 93.99 \\
80 & 93.92 & 93.98 & 94.08 & 94.11 \\
\midrule
\diagbox{\textbf{Training}}{\textbf{Testing}} & 
$\alpha =$ 1/255 & 1.5/255 & 2/255 & 2.5/255\\
\midrule
$\alpha = $ 1/255 & 95.07 & 91.31 & 89.33 & 88.95 \\
1.5/255 & 93.41 & 95.85 & 95.54 & 95.18 \\
2/255 & 91.51 & 95.88 & 96.09 & 96.02 \\
2.5/255 & 90.26 & 95.61 & 96.09 & 96.17 \\
\midrule
\bottomrule[1pt]
\end{tabular}%
}
\end{table}

For attacks like PGD, while hyperparameters  (step count $k$ and step size $\alpha$) exist, their influence on model parsing is less notable compared to the attack strength $\epsilon$ (Fig.\,\ref{fig: mismatch_attack_strength_selected}). \textbf{Table\,\ref{tab:PGDstep}} shows additional justification of model parsing vs. $k$ and $\alpha$.

\begin{table}[htbp]
\centering
\caption{Model parsing accuracy of MI/DMI-FGSM attacks vs. PGD $\ell_\infty$-attack on (CIFAR10, ResNet9), expanded from Table \ref{tab: in_distribution_arch_selected}.}
\label{tab: IFGSM}
\resizebox{0.35\textwidth}{!}{%
\begin{tabular}{c|c|c|c|c}
\toprule[1pt]
\midrule
{Strength} & {Attack} & \multicolumn{3}{c}{Accuracy (\%)} \\
($\epsilon$) & methods & $\mathbf{x^\prime}$ & $\boldsymbol{\delta}_{\mathrm{PEN}}$ & $\boldsymbol \delta$\\
\midrule
\multirow{3}{*}{8/255} & PGD & 66.62 & 83.20 & 95.07 \\
& MI & 65.16 & 82.46 & 94.33 \\
& DMI & 62.65 & 81.90 & 90.93 \\
\midrule
\multirow{3}{*}{12/255} & PGD & 76.65 & 89.73 & 94.91 \\
& MI & 72.83 & 87.02 & 93.86 \\
& DMI & 71.99 & 85.05 & 90.67 \\
\midrule
\bottomrule[1pt]
\end{tabular}%
}
\end{table}

\textbf{Tab.\,\ref{tab: IFGSM}} consistently shows that the improved transfer attacks like MI-FGSM \cite{dong2018boosting} and DMI-FGSM \cite{xie2019improving} are a bit harder in model parsing than the ordinary PGD attacks. However, the model parsing ability is still prominent, proving the feasibility of our method.